\documentclass[usenatbib, onecolumn]{rasti}

\usepackage{amssymb}
\usepackage{newtxtext,newtxmath}
\usepackage[T1]{fontenc}

\DeclareRobustCommand{\VAN}[3]{#2}
\let\VANthebibliography\thebibliography
\def\thebibliography{\DeclareRobustCommand{\VAN}[3]{##3}\VANthebibliography}

\usepackage{amsmath}
\usepackage{natbib}
\usepackage{physics}
\usepackage{bm}
\usepackage{graphicx}

\DeclareMathOperator*{\argmin}{\arg\!\min}
\newcommand{\tran}{\mathrm{t}}

\title[Extending Occam's inversion with lasso fusion]{Extending Occam's inversion with lasso fusion, overcomplete dictionaries, and isotropic total variation regularisation}

\author[A. Ray]{
Anandaroop Ray\thanks{E-mail: anandaroop.ray@ga.gov.au}
\\
Geoscience Australia, Australian Capital Territory, Australia
}

\date{Accepted XXX. Received YYY; in original form ZZZ}

\pubyear{\the\year{}}

\begin{document}
\label{firstpage}
\pagerange{\pageref{firstpage}--\pageref{lastpage}}
\maketitle

% Abstract of the paper
\begin{abstract}
Occam's inversion is a robust algorithm to perform nonlinear geophysical inversion. It provides the smoothest model within observation noise, thereby discouraging geological overinterpretation. While Occam originally penalised $\ell_2$ model roughness, $\ell_1$ can be used to provide models that are visually sharp. However, $\ell_1$ regularised geophysical inversion has diverged from the larger body of statistics and imaging literature. For example, $\ell_1$ regularisation with a difference operator (i.e., total variation) in multiple dimensions has two distinct forms, only one of which is invariant to edge orientation -- a distinction often overlooked in geophysics. In one dimension this reduces to the fused lasso problem in statistics. Within the Occam framework, for $\ell_2$ data norm and $\ell_1$ model regularisation, we show that lasso fusion through either synthesis or analysis leads to the same 1D problem. We extend the framework to multiple dimensions and to operators such as wavelet transforms in a dictionary, unravelling the mathematics behind three commonly used solvers for $\ell_1$ regularised problems. These are coordinate descent, iteratively reweighted least squares (IRLS), and split Bregman. We use them to solve a sequence of problems that are linear (1D regression, 2D deblurring) and nonlinear (1D airborne transient electromagnetics) including a field data example. We recommend coordinate descent for 1D problems, and IRLS over split Bregman for 2D problems, with IRLS requiring up to an order of magnitude fewer least squares solves. We hope this work will further encourage geophysicists to adopt $\ell_1$ regularised inversion, by presenting its various forms as a familiar Occam's inversion.
\end{abstract}

% Include between one and six keywords.

%%%%%%%%%%%%%%%%%%%%%%%%%%%%%%%%%%%%%%%%%%%%%%%%%%

%%%%%%%%%%%%%%%%% BODY OF PAPER %%%%%%%%%%%%%%%%%%

\section{Introduction}
The original Occam's inversion algorithm \citep{Constable1987} is a robust, nonlinear gradient descent algorithm extensively used in geophysical investigation, particularly for imaging subsurface electromagnetic conductivity. This conductivity is inverted from surface observations that are forward modelled with quasi-static induction physics. However, though Occam can be understood as belonging to the Gauss-Newton family, it distinguishes itself in a rather unique way. Initially, during the misfit reduction stage (phase I), a line search is performed to find the lowest misfit model. At a later stage when the line search provides a $\chi^2$ data misfit lower than the target noise (phase II), the algorithm deliberately seeks the smoothest model obtained through root-finding, such that the data misfit equals the known target noise level. This is why, when microprocessor clock-speeds were slow, and computer memory was not plentiful compared to today's standards, it was referred to as ``The Rolls-Royce of inversion algorithms" within the geophysical community (personal communication, Constable). However, as of this writing, over the last four decades, it has garnered over 4,000 citations and has reliably and extensively been used across a variety of geophysical fields from tectonophysics to hydrogeophysics \citep[e.g.,][to name only a few citing articles]{degroot1990occam, mckenzie1991partial, merrill1998magnetic, kelbert2014modem, binley2015emergence, Key2016, aster2018parameter, sarker20262d}. The method has provided a robust scheme for convergence to within noise, and has been improved upon over the years with a fast-Occam approach \citep{Brodie2010, Key2016} that shortens the line search. Occam's inversion provides a bulwark against geological over-interpretation, which was rife in the years before its advent \cite[see Chapter 6 of][for a modern exposition]{Wheelock2012b}. Curiously, this has a Bayesian character, though no Bayesian inference is necessary to formulate the algorithm. The ``Occam factor'' \citep[see][Chapter 28]{mackay2003information} or ``Ockham factor'' \citep{Malinverno2002, Ray2016} is the Bayesian analogue, which lays down the Bayesian framework for a ``simplest-but-no-simpler'' approach to statistical model selection within an inverse framework.

However, the idea of simplicity and $\ell_2$ smoothness were interchangeable in the original implementation. With few exceptions \citep[e.g.,][]{Farquharson1998}, this meant using an $\ell_2$ norm on the terms with a regularisation operator (e.g., first differences), as will be shown in the next section. \cite{vargas2025rambo} recently addressed this issue, with an $\ell_1$ regularised Occam's inversion, that allows visually sharp and sudden changes to be produced by a first differences regularisation operator. They specifically use a split Bregman method \citep{goldstein2009split}, and state that their results require a large number of inner least squares solves, even for the 1D problems they tried. They acknowledge that this makes it difficult to implement their approach for higher dimensional problems, which they have left as future work, providing proof of concept with a 1D magnetotelluric (MT) and 1D direct current (DC) resistivity example. In our work we investigate why split Bregman requires such a large number of least squares solves, and provide two alternatives. We use coordinate descent \citep{friedman2010regularization} for 1D problems, and demonstrate scalable implementations for 2D problems with both split Bregman and iteratively reweighted least squares \citep[IRLS, see e.g.,][]{green1984iteratively, daubechies2010iteratively}, with IRLS providing identical results to split Bregman at far lower computational cost. Further, while \cite{vargas2025rambo} use a first differences regularisation operator, we extend the Occam philosophy to a general framework that utilises a range of invertible and non-invertible regularisation operators, allowing the full machinery of $\ell_1$ regularisation with dictionary synthesis as well as analysis \citep{elad2007analysis} to be used within a principled framework that allows for the selection of model complexity in accordance with the original Occam \citep{Constable1987} paper. Furthermore, our interest in this work is not on uncertainty estimation. However, we show that the modifications to IRLS we propose require an order of magnitude fewer least squares solves for the same nonlinear total variation problem, than it does with split Bregman. This could make the uncertainty estimation approach described in \cite{vargas2025rambo}, tractable for nonlinear 2D and 3D problems. Leaving aside the question of uncertainty, without explicitly forming the normal equations, we could potentially make the Occam method competitive in 3D, for $\ell_1$ (as well as $\ell_2$) problems. We show this with both IRLS and split Bregman in 2D, using Golub-Kahan bidiagonalised least squares solvers \citep{fong2011lsmr}. This idea is similar, but not the same as illustrated in~\cite{egbert2012hybrid} for $\ell_2$ regularisation, which uses a covariance for regularisation and operates in the data space \citep[see][]{siripunvaraporn2005three}.
 
As described by \cite{donoho1989uncertainty}, there is (perhaps surprisingly) a long history of $\ell_1$ regularisation in geophysical inversion. Early efforts were in seismic deconvolution \citep[e.g.,][]{levy1981reconstruction, oldenburg1983recovery, walker1983autoregressive, santosa1986linear}. Extension to the nonlinear electromagnetic inverse problem was made not long afterwards \citep[e.g.,][]{dosso1990inversion, oldenburg1993efficient, ellis1993approximate, Farquharson1998}. More recently, there has been much work across various geophysical fields, ranging from seismic imaging \citep[e.g.,][]{hennenfent2008simply, herrmann2012efficient}, seismic full waveform inversion \citep[e.g.,][]{yong2018total, esser2018total}, geophysical tomography \citep{Loris2010, turuncctur2023overcomplete} to inductive electromagnetics \citep[e.g.,][]{farquharson2008constructing, kangazian2026constructing}, to cite only a few examples. Importantly, however, as pointed out by \cite{sun2024understanding}, implementing total variation in 2D or 3D is not simply an extension of the $\ell_1$ norm to difference operators in multiple dimensions. What naturally falls out by doing so, is anisotropic total variation, which prefers axis aligned changes. This is not desirable, and it seems there is some confusion in the geophysics community when difference regularisation in $\ell_1$ is invoked, as it generally seems to imply the anisotropic version. We derive both the anisotropic and the orientation invariant isotropic version in our treatment and illustrate the difference.

As alluded to in \cite{vargas2025rambo}, the adoption of $\ell_1$ norm regularisation by the geophysics community, however, has not been rapid or widespread. Indeed, the related literature spanning optimisation and statistics that is directly applicable to geophysical inversion is vast and highly specialised \citep[e.g.,][]{tibshirani1996regression, candes2006robust, esser2010general, boyd2011distributed}. Coming to terms with differing mathematical descriptions, and answering questions of computational tractability between various approaches to solve what ultimately amounts to the same problem, is a difficult task for the majority of geophysicists, be they seasoned practitioners or advanced researchers. In this work, we bring together the synthesis and analysis forms \citep[see e.g.,][]{candes2011compressed} of the $\ell_1$ regularised problem within a unified framework, view it in terms of the fused lasso \citep{tibshirani2005sparsity}, generalise it to higher dimensions, compare various implementation strategies for both linear and nonlinear problems, and present it as a familiar Occam's inversion. Importantly, we do not avoid a minimum of mathematics, and work through the derivations that can be traced only through a multitude of references. 

\section{Methodology}
The regularised inverse problem can be written as a minimisation of the following objective function over $\mathbf m$:
\begin{equation}
    \argmin_\mathbf{m} \phi(\mathbf{m}) = \norm{\mathbf{W}(\mathbf{d}-\mathbf{f}(\mathbf{m}))}^2 + \lambda^2 \left[\norm{\mathbf{Rm}}^p_p + \beta^2 \norm{\mathbf{m} - \mathbf{m_0}}^p_p\right], \label{eqn:phi}
\end{equation}
where $\mathbf{m}$ is the conductivity model, $\mathbf{f}(\cdot)$ is the forward operator, $\mathbf{d}$ contains observed data, and $\mathbf{W}$ are the data weights that relate to the data covariance matrix $\mathbf{C_d}$ such that $\mathbf{C_d^{-1}} = \mathbf{W^tW}$. $\mathbf{R}$ is a first differences regularisation operator, $\lambda^2 >0$ is the regularisation parameter, $\mathbf{m_0}$ is a reference model, and $\beta^2$ is a positive scalar that dictates bias towards the reference model. Based on the Occam's inversion algorithm, we linearise the problem by adopting a Gauss-Newton approach to minimise the objective function. This involves sweeping through $\lambda^2$ to identify the largest (and therefore smoothest) $\lambda^2$ value that fits the model within the accepted data noise. Crucially, the original \cite{Constable1987} paper used the regularisation norm $p=2$. This choice corresponds to the traditional $\ell_2$ norm on the regularisation terms, corresponding loosely to a Normal Bayesian prior. It is also mathematically differentiable everywhere, making it a natural and attractive choice. However, it is known to provide visually smooth and oscillatory solutions as demonstrated in this work \citep[see also][]{loris2007tomographic}. 

In our implementation we instead set $p=1$ and use the idea of subgradients (see Appendix~\ref{sec:whycd}) that get around the non-differentiability of the absolute value function by solving the \textit{fused lasso} problem \citep{tibshirani2005sparsity}. Loosely speaking, this choice of $p$ corresponds to a Laplace Bayesian prior. The fused lasso aims to keep the model smooth in an $\ell_1$ sense, by setting unnecessary first difference elements to zero, while also setting deviations from the reference model close to zero. One way to understand how this works, is to recall that a Laplace distribution has heavier tails and a sharper peak than a Normal distribution. In the Bayesian prior analogy to \eqref{eqn:phi}, realisations correspond to first differences $\mathbf{Rm}$ and deviations from the reference model $\mathbf{m} - \mathbf{m_0}$. Therefore with the $\ell_1$ norm, compared to the $\ell_2$ norm, some large values are allowed, while simultaneously concentrating the majority of values near zero. For a more mathematical explanation invoking $\ell_p$ balls, see \cite{donoho2006compressed}. We then extend the fused-lasso formulation with operators beyond difference matrices with families of basis functions such as discrete wavelet and cosine transforms \citep{daubechies1992ten, elad2007analysis}. We use dictionary synthesis to develop and arrive at the rationale for total variation (TV) forms of the regularisation operator in multiple dimensions, demonstrating scalable implementation using both iteratively reweighted least squares \citep{hunter2004tutorial, daubechies2010iteratively} and the split Bregman method \citep{goldstein2009split}. 
\section{Linear fused-lasso formulation}
We first consider the linear inverse problem
\begin{equation}
    \mathbf{y} = \mathbf{A}\mathbf{m},
    \label{eq:linear_forward}
\end{equation}
where $\mathbf{y}\in\mathbb{R}^{n}$ is the data vector, $\mathbf{m}\in\mathbb{R}^{p}$ is the model vector, and $\mathbf{A}\in\mathbb{R}^{n\times p}$ is the forward matrix. Let $\mathbf{m}_0\in\mathbb{R}^{p}$ denote a reference model, and define the shifted model variable
\begin{equation}
    \mathbf{m}' = \mathbf{m} - \mathbf{m}_0 .
    \label{eq:shifted_model}
\end{equation}
The corresponding shifted data vector is
\begin{equation}
    \mathbf{y}' = \mathbf{y} - \mathbf{A}\mathbf{m}_0 ,
    \label{eq:shifted_data}
\end{equation}
so that the linear inverse problem may be written in terms of
$\mathbf{m}'$ as
\begin{equation}
    \mathbf{y}' = \mathbf{A}\mathbf{m}' .
    \label{eq:shifted_forward}
\end{equation}
Modifying the data misfit term in \eqref{eqn:phi} slightly to be more in line with the $\ell_1$ literature, we consider the regularised objective function 
\begin{equation}
    \phi(\mathbf{m}')
    =
    \frac{1}{2}
    \left\|
        \mathbf{y}' - \mathbf{A}\mathbf{m}'
    \right\|_2^2
    +
    \lambda^2
    \left[
        \beta^2
        \left\|
            \mathbf{m}'
        \right\|_1
        +
        \left\|
            \mathbf{D}\mathbf{m}'
        \right\|_1
    \right],
    \label{eq:fused_lasso_shifted}
\end{equation}
where $\lambda^2>0$ controls the overall strength of regularisation and $\beta^2\geq 0$ controls shrinkage towards the reference model. The first regularisation term promotes sparsity in $\mathbf m'$. In the absence of the second regularisation term, \eqref{eq:fused_lasso_shifted} is the standard lasso problem \citep{tibshirani1996regression}. Lasso, as it is commonly written, is the `least absolute shrinkage selection operator'. When $\mathbf{D}$ is chosen as a first-difference operator, minimising \eqref{eq:fused_lasso_shifted} is known as the fused lasso problem of \citet{tibshirani2005sparsity}. This is because the second term in the regularisation is a one-dimensional total-variation or fusion penalty, closely related to the total-variation regularisation introduced by \citet{rudin1992nonlinear}. The fusion penalty promotes sparsity in the first differences of $\mathbf m'$, analogous to role of the first difference smoothness penalty with $\ell_2$ regularisation.

For the present formulation, we take the reference model $\mathbf{m}_0$ to be constant. This choice keeps the interpretation of the fusion penalty simple: the first-difference part of $\mathbf{D}(\mathbf{m}-\mathbf{m}_0)$ is then equivalent to the first differences operation $\mathbf{Dm}$ itself. Thus the regularisation promotes sparsity in model jumps rather than sparsity in deviations from a prior model with structure. A non-constant reference model is mathematically permissible, but in that case the fusion penalty encourages the recovered model to preserve the differences already present in the reference model, which is a different prior assumption. In this work we use the anchored first-difference operator
\begin{equation}
    \mathbf{D}
    =
    \begin{bmatrix}
        \eta & 0 & 0 & \cdots & 0 & 0 \\
        -1 & 1 & 0 & \cdots & 0 & 0 \\
        0 & -1 & 1 & \cdots & 0 & 0 \\
        \vdots & \vdots & \vdots & \ddots & \vdots & \vdots \\
        0 & 0 & 0 & \cdots & -1 & 1
    \end{bmatrix}.
    \label{eq:D_operator}
\end{equation}
Here $\eta\geq 0$ controls the strength of the `anchor' on the first component of $\mathbf m'$, while the remaining rows compute first differences between adjacent components. Thus
\begin{equation}
    \mathbf{D}\mathbf{m}'
    =
    \begin{bmatrix}
        \eta m'_1 \\
        m'_2 - m'_1 \\
        m'_3 - m'_2 \\
        \vdots \\
        m'_p - m'_{p-1}
    \end{bmatrix},
    \label{eq:Dx_expanded}
\end{equation}
and therefore
\begin{equation}
    \left\|\mathbf{D}\mathbf{m}'\right\|_1
    =
    \eta |m'_1|
    +
    \sum_{j=2}^{p}
    \left|m'_j-m'_{j-1}\right| .
    \label{eq:Dx_norm_expanded}
\end{equation}
The case $\eta=0$ gives exactly the usual one-dimensional fused-lasso fusion penalty. However, the resulting first-difference operator is singular and cannot be inverted. Conversely, taking $\eta=1$ gives an invertible operator, but introduces a full-strength penalty on the first element in $\mathbf m'$, $m'_1$, which is undesirable. We therefore choose $0<\eta\ll 1$. This gives a weakly anchored approximation to the fused-lasso penalty while preserving the invertibility needed for the transformation below. Since $\det(\mathbf{D})=\eta$, the matrix $\mathbf{D}$ is invertible for $\eta\neq 0$ \citep[see][example 1]{calvetti2020sparsity}. Let
\begin{equation}
    \mathbf{z} = \mathbf{D}\mathbf{m}',
    \qquad
    \mathbf{C} = \mathbf{D}^{-1}.
    \label{eq:z_transform}
\end{equation}
Then
\begin{equation}
    \mathbf{m}' = \mathbf{C}\mathbf{z},
    \label{eq:xprime_Cz}
\end{equation}
and hence
\begin{equation}
    \mathbf{m} = \mathbf{m}_0 + \mathbf{C}\mathbf{z}.
    \label{eq:x_from_z}
\end{equation}
The inverse matrix $\mathbf{C}$ is a cumulative-sum operator,
\begin{equation}
    \mathbf{C}
    =
    \begin{bmatrix}
        \eta^{-1} & 0 & 0 & \cdots & 0 \\
        \eta^{-1} & 1 & 0 & \cdots & 0 \\
        \eta^{-1} & 1 & 1 & \cdots & 0 \\
        \vdots & \vdots & \vdots & \ddots & \vdots \\
        \eta^{-1} & 1 & 1 & \cdots & 1
    \end{bmatrix}.
    \label{eq:C_operator}
\end{equation}
Defining
\begin{equation}
    \mathbf{T} = \mathbf{A}\mathbf{C},
    \label{eq:T_operator}
\end{equation}
the objective function in terms of $\mathbf{z}$ becomes
\begin{equation}
    \phi(\mathbf{z})
    =
    \frac{1}{2}
    \left\|
        \mathbf{y}' - \mathbf{T}\mathbf{z}
    \right\|_2^2
    +
    \lambda^2
    \left[
        \beta^2
        \left\|
            \mathbf{C}\mathbf{z}
        \right\|_1
        +
        \left\|
            \mathbf{z}
        \right\|_1
    \right].
    \label{eq:z_objective_full}
\end{equation}
This transformation is useful because the approximate fusion penalty becomes the ordinary lasso penalty $\|\mathbf{z}\|_1$. The vector $\mathbf{z}$ contains a weakly anchored first component followed by successive jumps in $\mathbf{m'}$.  Consequently, sparsity in $\mathbf{z}$ corresponds to an $\mathbf{m'}$ that is approximately piecewise constant -- i.e., components of $\mathbf m'$ are ``fused'' together.

We now follow the pathwise strategy used for fused-lasso problems by \citet{friedman2007}: first solve the problem with the coefficient-shrinkage penalty set to zero, and then introduce that penalty through soft thresholding. In our notation, the coefficient-shrinkage penalty is controlled by $\lambda^2\beta^2$. Thus, setting $\beta^2=0$ gives the pure fusion-penalised problem. Under the invertible approximation above, equation~\eqref{eq:z_objective_full} then reduces to
\begin{equation}
    \hat{\mathbf{z}}
    =
    \arg\min_{\mathbf{z}}
    \left\{
        \frac{1}{2}
        \left\|
            \mathbf{y}' - \mathbf{T}\mathbf{z}
        \right\|_2^2
        +
        \lambda^2
        \left\|
            \mathbf{z}
        \right\|_1
    \right\}.
    \label{eq:z_lasso_problem}
\end{equation}
Thus, the pure fusion-regularised problem in $\mathbf{m}'$ is transformed into a standard lasso problem in $\mathbf{z}$. This problem may be solved using any efficient lasso solver, including coordinate-descent methods \citep{friedman2010regularization}. An explanation of the coordinate descent, shrinkage, and shrinkage of coefficients to exact zeros is provided in Appendix~\ref{sec:whycd}. The estimate of $\mathbf{m'}$ so obtained is $\mathbf{C}\hat{\mathbf{z}}$, and the corresponding model estimate is
\begin{equation}
    \hat{\mathbf{m}}
    =
    \mathbf{m}_0 + \mathbf{C}\hat{\mathbf{z}}.
    \label{eq:xhat_from_zhat}
\end{equation}
For $\beta^2>0$, the additional term $\beta^2\|\mathbf{m}'\|_1$ encourages $\mathbf{m'}$ itself to be sparse, i.e., deviations of $\mathbf m$ from the reference model $\mathbf{m}_0$ become sparse. Motivated by the soft-thresholding result used in pathwise fused-lasso algorithms \citep{friedman2007}, we apply componentwise soft thresholding to
$\mathbf{C}\hat{\mathbf{z}}$. Let
\begin{equation}
    S(u,\tau)
    =
    \operatorname{sign}(u)\max\{|u|-\tau,0\}
    \label{eq:soft_threshold_scalar}
\end{equation}
denote the scalar soft-thresholding operator \citep{donoho1995noising}, applied componentwise to vectors.
We then define the soft-thresholded (ST) estimate
\begin{equation}
    \mathbf{m}_{\text{ST}}
    =
    \mathbf{m}_0
    +
    S
    \left(
        \mathbf{C}\hat{\mathbf{z}},
        \lambda^2\beta^2
    \right).
    \label{eq:double_hat_estimate}
\end{equation}
This construction preserves the approximately piecewise-constant character promoted by the fusion penalty while additionally shrinking $\mathbf{m-m_0}$ towards zero. The soft-thresholding step~\eqref{eq:double_hat_estimate} is exact for the fused-lasso signal approximator, under the condition $\mathbf{A}=\mathbf{I}$, as described by \citet[][see also our \eqref{eq:lasso_componentwise_independent} in Appendix~\ref{sec:whycd}]{friedman2007}, but is not guaranteed to solve the general fused-lasso regression problem for arbitrary $\mathbf{A}$. In the present work, following the strategy suggested by \citet{friedman2007} for solving the general fused-lasso, we use the same idea: first solve the $\beta^2=0$ weakly anchored fusion problem exactly as a lasso problem in $\mathbf{z}$, and then apply the soft-thresholding step to incorporate the $\beta^2\|\mathbf{m'}\|_1$ shrinkage term. This two-step procedure provides our practical solution strategy for the weakly anchored fused-lasso formulation with $0<\eta \ll1$. To be clear, our  approximation only affects shrinkage or closeness to the reference model $\mathbf m_0$ and not the fusion part of the problem. To be explicit, the part ensuring $\ell_1$ smoothness in the solution (i.e., fusion) is decoupled from the reference shrinkage approximation -- and it is the fusion term that is of primary importance in most geophysical inversion problems. Further, we will show in Section~\ref{sec:weighting} that we can free ourselves from selection of the parameter $\eta$. 

\section{Generalising the transform operator}
The preceding derivation was written for an anchored first-difference operator $\mathbf{D}$, motivated by the one-dimensional fused lasso. However, the reparameterisation itself is more general. The only algebraic requirement is that $\mathbf{D}\in\mathbb{R}^{p\times p}$ be invertible, so that
\begin{equation}
    \mathbf{z} = \mathbf{D}\mathbf{m}',
    \qquad
    \mathbf{m}' = \mathbf{D}^{-1}\mathbf{z}.
    \label{eq:general_D_transform}
\end{equation}
Writing
\begin{equation}
    \mathbf{C} = \mathbf{D}^{-1},
    \qquad
    \mathbf{T} = \mathbf{A}\mathbf{C},
    \label{eq:general_C_T}
\end{equation}
the case $\beta^2=0$ becomes
\begin{equation}
    \hat{\mathbf{z}}
    =
    \arg\min_{\mathbf{z}}
    \left\{
        \frac{1}{2}
        \left\|
            \mathbf{y}'-\mathbf{T}\mathbf{z}
        \right\|_2^2
        +
        \lambda^2
        \left\|
            \mathbf{z}
        \right\|_1
    \right\}.
    \label{eq:general_D_lasso}
\end{equation}
Thus, any invertible transform $\mathbf{D}$ converts an $L_1$ penalty on $\mathbf{D}\mathbf{m}'$ into an ordinary lasso penalty on the transformed coefficients $\mathbf{z}$. The interpretation of the regularisation is then determined by the choice of $\mathbf{D}$. For the anchored first-difference operator, sparsity in $\mathbf{z}$ corresponds to sparsity in model jumps. For a wavelet transform operator. \citep{mallat1989theory}, sparsity in $\mathbf{z}$ corresponds to sparsity in wavelet coefficients. This observation also provides a natural route to higher-dimensional extensions. Let
\begin{equation}
    \mathbf{M}'\in\mathbb{R}^{n_{p_1}\times n_{p_2}}
\end{equation}
denote a two-dimensional shifted model (henceforth referred to as perturbation), and let
\begin{equation}
    \mathbf{m}' = \operatorname{vec}(\mathbf{M}')
\end{equation}
be its column-major vectorisation. Suppose
\begin{equation}
    \mathbf{D}_1\in\mathbb{R}^{n_{p_1}\times n_{p_1}},
    \qquad
    \mathbf{D}_2\in\mathbb{R}^{n_{p_2}\times n_{p_2}}
\end{equation}
are invertible one-dimensional transforms acting along the first and second dimensions, respectively. Applying these transforms separably gives
\begin{equation}
    \mathbf{Z}
    =
    \mathbf{D}_1 \mathbf{M}' \mathbf{D}_2^{\tran}.
    \label{eq:separable_matrix_transform}
\end{equation}
Using the identity for column-major vectorisation
\begin{equation}
    \operatorname{vec}
    \left(
        \mathbf{B}\mathbf{M}\mathbf{A}^{\tran}
    \right)
    =
    \left(
        \mathbf{A}\otimes\mathbf{B}
    \right)
    \operatorname{vec}(\mathbf{M}),
    \label{eq:vec_identity}
\end{equation}
equation~\eqref{eq:separable_matrix_transform} may be written as
\begin{equation}
    \mathbf{z}
    =
    \mathbf{D}_{2D}\mathbf{m}',
    \qquad
    \mathbf{D}_{2D}
    =
    \mathbf{D}_2\otimes\mathbf{D}_1.
    \label{eq:2D_kron_D}
\end{equation}
If both $\mathbf{D}_1$ and $\mathbf{D}_2$ are invertible, then
$\mathbf{D}_{2D}$ is also invertible, with
\begin{equation}
    \mathbf{D}_{2D}^{-1}
    =
    \mathbf{D}_2^{-1}\otimes\mathbf{D}_1^{-1}.
    \label{eq:2D_kron_inverse}
\end{equation}
Therefore, the same lasso reparameterisation applies in two dimensions, with
\begin{equation}
    \mathbf{T}
    =
    \mathbf{A}\mathbf{D}_{2D}^{-1}.
    \label{eq:2D_transformed_forward}
\end{equation}
The construction extends similarly to three dimensions. It is important, however, to distinguish separable invertible transforms from standard multidimensional total variation (TV). In two dimensions, anisotropic TV usually penalises the stacked directional gradient,
\begin{equation}
    \left\|
        \left(
            \mathbf{I}_{n_{p_2}}\otimes\mathbf{D}_1
        \right)
        \mathbf{m}'
    \right\|_1
    +
    \left\|
        \left(
            \mathbf{D}_2\otimes\mathbf{I}_{n_{p_1}}
        \right)
        \mathbf{m}'
    \right\|_1.
    \label{eq:standard_2D_TV}
\end{equation}
The above operator (to be discussed in detail in Section~\ref{sec:tvall}) is generally rectangular and is not directly invertible. By contrast, the separable operator
\begin{equation}
    \mathbf{D}_{2D}
    =
    \mathbf{D}_2\otimes\mathbf{D}_1 \label{eq:mixed}
\end{equation}
is square and invertible when its one-dimensional factors are invertible. If $\mathbf{D}_1$ and $\mathbf{D}_2$ are difference-like operators, then the $\ell_1$ norm of \eqref{eq:mixed} gives a mixed difference sharpness-promoting transform that is related but not identical to standard multidimensional TV.

Other choices of $\mathbf{D}$ give different sparsity assumptions. For example, if $\mathbf{D}$ is chosen to be an orthogonal discrete wavelet transform \citep[e.g.,][]{daubechies1992ten}, then $\mathbf{D}^{-1}=\mathbf{D}^{\tran}$ and $\mathbf{z}$ contains the wavelet coefficients of the model perturbation. In that case equation~\eqref{eq:general_D_lasso} is a lasso problem in wavelet coefficients. Using the Haar basis gives a particularly simple example: sparsity in the Haar coefficients promotes a piecewise-constant, multiscale representation of $\mathbf{m}'$ \citep[e.g.,][]{Hawkins2015, Ray2018}. Unlike the anchored difference operator, an orthogonal Haar transform is an exact invertible transform rather than an approximation to a singular TV operator.

Thus, the anchored first-difference formulation used above is one member of a broader class of sparse-transform inverse problems. The common structure is to choose an invertible transform $\mathbf{D}$ whose coefficients have a useful geophysical interpretation, solve the corresponding lasso problem in $\mathbf{z}$, and then recover the model perturbation through $\mathbf{m}'=\mathbf{D}^{-1}\mathbf{z}$.

\section{Weighting the transformed coefficients}
\label{sec:weighting}
Motivated by the fact that lower levels of wavelet transform coefficients concentrate coarser features, we can introduce an explicit weighting that penalises higher-level coefficients more strongly. Rather than altering the transform, we attach a diagonal penalty weight $\bm\Lambda$ directly to the $\ell_1$ regularisation term. Keeping the transform and its inverse unweighted,
\begin{equation}
    \mathbf z = (\mathbf{D}_2 \otimes \mathbf{D}_1)\, \mathbf{m}',
    \qquad
    \mathbf{T} = \mathbf{A}\,(\mathbf{D}_2^{-1} \otimes \mathbf{D}_1^{-1}),
    \label{eq:weighted_forward}
\end{equation}
the regularisation becomes the weighted lasso penalty $\|\bm\Lambda \mathbf z\|_1$, where for a separable transform
$\bm\Lambda = \bm\Lambda_2 \otimes \bm\Lambda_1$ assigns the weight $\Lambda_{2,j}\,\Lambda_{1,i}$ to the coefficient indexed $(i,j)$. A coefficient with weight one is penalised at the shared strength $\lambda^2$. A weight greater than one suppresses it more strongly, and a weight of zero leaves it free to be fit by the data without shrinkage (used also for the model level and for wavelet scaling coefficients). Explicitly, \eqref{eq:z_objective_full} is modified to the following  objective function
\begin{equation}
    \phi(\mathbf{z})
    =
    \frac{1}{2}
    \left\|
        \mathbf{y}' - \mathbf{T}\mathbf{z}
    \right\|_2^2
    +
    \lambda^2
    \left[
        \beta^2
        \left\|
            \mathbf{C}\mathbf{z}
        \right\|_1
        +
        \left\|
            \bm\Lambda\mathbf{z}
        \right\|_1
    \right].
    \label{eq:z_objective_full_weighted}
\end{equation}
At this juncture, the weighted construction clarifies a limitation of separable transform coefficient weighting. If we try to penalise first differences in one direction more than the other by setting the directional weight $\bm\Lambda_2$ to a scalar $>1$ and $\bm\Lambda_1=\mathbf{I}$, the scalar factors out of the separable penalty $\|(\bm\Lambda_2\otimes\bm\Lambda_1)\mathbf{z}\|_1$ and merely rescales $\lambda^2$, leaving the relative regularisation between the two directions unchanged. Genuinely separate ``difference strengths'' in each direction therefore require the approaches discussed later in Section~\ref{sec:tvall}. However, a useful consequence of \eqref{eq:z_objective_full_weighted} is that setting $\Lambda_1 = 0$ immediately allows us to free the first difference coefficient $(\mathbf z)_1$ from the regularisation. It is then freely minimised only through minimisation of the data norm. Therefore, we set $\eta$ in \eqref{eq:C_operator} to 1 for a variety of problems that require an invertible regularisation operator. In certain cases, we explicitly work with a difference operator $\mathbf D$ that has $\eta$ set to zero, or a cumulative sum integrator $\mathbf C$ that has $\frac 1 \eta$ set to zero. This will be clear in the synthesis formulation to be discussed next.

\section{Synthesis formulation: the linear problem}
\label{sec:synthesis_linear}

The formulations above are invertible, with a synthesis map $\mathbf{m}'=\mathbf{C}\mathbf{z}$. The requirement that $\mathbf{D} = \mathbf{C}^{-1}$ be square and invertible is not necessary. We can more generally pose the problem directly in synthesis form. Rather than seeking a model whose coefficients are sparse, we build the model as a sparse superposition of atoms drawn from a dictionary of synthesis operators \citep{elad2007analysis, turuncctur2023overcomplete}. This dictionary contains more bases than are necessary to represent the model and is regarded as being overcomplete. Let $\mathbf{G}_i\in\mathbb{R}^{n_p\times n_{p_i}}$, $i=1,\dots,K$, be a collection of synthesis operators, each mapping a coefficient vector to model space. Examples include inverse-difference (cumulative-sum) operators acting along a single direction, discrete cosine transforms, and inverse wavelet transforms of one or more families. We collect the atoms, unweighted, into the dictionary
\begin{equation}
    \mathbf{C}
    =
    \begin{bmatrix}
        \mathbf{G}_1
        &
        \mathbf{G}_2
        &
        \cdots
        &
        \mathbf{G}_K
    \end{bmatrix}
    \in\mathbb{R}^{n_p\times P},
    \qquad
    P = \sum_{i=1}^{K} n_{p_i},
    \label{eq:synthesis_dictionary}
\end{equation}
with the corresponding stacked coefficient vector
\begin{equation}
    \mathbf{z}
    =
    \begin{bmatrix}
        \mathbf{z}_1 \\ \mathbf{z}_2 \\ \vdots \\ \mathbf{z}_K
    \end{bmatrix}
    \in\mathbb{R}^{P},
    \label{eq:synthesis_coeffs}
\end{equation}
so that the model perturbation is synthesised as
\begin{equation}
    \mathbf{m}'
    =
    \mathbf{C}\mathbf{z}
    =
    \sum_{i=1}^{K}
    \mathbf{G}_i\mathbf{z}_i .
    \label{eq:synthesis_model}
\end{equation}
We deliberately reuse the symbol $\mathbf{C}$ from \eqref{eq:general_C_T}: it plays exactly the same role, mapping coefficients to the model. The difference is that $\mathbf{C}$ is no longer a square inverse $\mathbf{D}^{-1}$ but a generally rectangular, overcomplete dictionary, with no inverse. The single-basis invertible case is recovered as the special case $K=1$, $\mathbf{G}_1=\mathbf{D}^{-1}$, for which $\mathbf{C}$ is square and \eqref{eq:synthesis_model} reduces to $\mathbf{m}'=\mathbf{D}^{-1}\mathbf{z}$.

All prior preferences between and within the atoms are carried by the single diagonal penalty weight $\bm\Lambda$ of Section~\ref{sec:weighting}, now indexed over the full $P$-dimensional coefficient vector. With the dictionary fixed, the regularised objective over the coefficients is
\begin{equation}
    \phi(\mathbf{z})
    =
    \frac{1}{2}
    \left\|
        \mathbf{y}' - \mathbf{A}\mathbf{C}\mathbf{z}
    \right\|_2^2
    +
    \lambda^2
    \left[
        \beta^2
        \left\|
            \mathbf{C}\mathbf{z}
        \right\|_1
        +
        \left\|
            \bm\Lambda\mathbf{z}
        \right\|_1
    \right],
    \label{eq:synthesis_objective_full}
\end{equation}
where, as before, $\mathbf{y}'=\mathbf{y}-\mathbf{A}\mathbf{m}_0$ is the shifted data, $\lambda^2>0$ sets the overall regularisation strength, and $\beta^2\geq 0$ controls shrinkage of the recovered model perturbation towards zero, i.e.\ of the model towards the reference $\mathbf{m}_0$. As in Section~\ref{sec:weighting}, a coefficient with $\Lambda_j=1$ is penalised at the shared strength $\lambda^2$, one with $\Lambda_j>1$ is suppressed more strongly, and one with $\Lambda_j=0$ is removed from the fusion penalty entirely, free to take whatever value best fits the data without shrinkage. The fusion penalty $\|\mathbf{D}\mathbf{m}'\|_1$ of the invertible formulation has thus been replaced by the weighted lasso penalty $\|\bm\Lambda\mathbf{z}\|_1$ acting on the dictionary coefficients: sparsity in $\mathbf{z}$ now means that the model perturbation is explained by a small number of atoms, drawn from whichever bases represent the data most parsimoniously (in combination). Defining the forward matrix exactly as in \eqref{eq:general_C_T},
\begin{equation}
    \mathbf{T}
    =
    \mathbf{A}\mathbf{C}
    \in\mathbb{R}^{n_d\times P},
    \label{eq:synthesis_T}
\end{equation}
the case $\beta^2=0$ becomes the weighted lasso problem
\begin{equation}
    \hat{\mathbf{z}}
    =
    \arg\min_{\mathbf{z}}
    \left\{
        \frac{1}{2}
        \left\|
            \mathbf{y}' - \mathbf{T}\mathbf{z}
        \right\|_2^2
        +
        \lambda^2
        \left\|
            \bm\Lambda\mathbf{z}
        \right\|_1
    \right\},
    \label{eq:synthesis_lasso_problem}
\end{equation}
of the same form as \eqref{eq:general_D_lasso}, but with $\mathbf{z}$ now existing in the $P$-dimensional coefficient space of the dictionary rather than in the $p$-dimensional model space. The corresponding model estimate is
\begin{equation}
    \hat{\mathbf{m}}
    =
    \mathbf{m}_0
    +
    \mathbf{C}\hat{\mathbf{z}},
    \label{eq:synthesis_xhat}
\end{equation}
and, for $\beta^2>0$, the soft-thresholded estimate carries over unchanged from
\eqref{eq:double_hat_estimate},
\begin{equation}
    \mathbf{m}_{\mathrm{ST}}
    =
    \mathbf{m}_0
    +
    S
    \left(
        \mathbf{C}\hat{\mathbf{z}},
        \lambda^2\beta^2
    \right),
    \label{eq:synthesis_xST}
\end{equation}
with $S$ the componentwise soft-thresholding operator of
\eqref{eq:soft_threshold_scalar}.

\section{Total variation regularisation}\label{sec:tvall}
\subsection{Anisotropic total variation}
\label{sec:synthesis_TV}
In more than one dimension the dictionary makes the relationship to total variation explicit. First, we can recall from \eqref{eq:standard_2D_TV} that genuine anisotropic total variation penalises the directional differences separately,
$\|(\mathbf{I}_{n_{p_2}}\otimes\mathbf{D}_1)\mathbf{m}'\|_1
+\|(\mathbf{D}_2\otimes\mathbf{I}_{n_{p_1}})\mathbf{m}'\|_1$. This is a sum of two penalties that do not cancel against one another. For each pixel $p$,  with $\mathbf D_y = \mathbf I_{n_{p_2}} \otimes \mathbf D_1$, and $\mathbf D_x = \mathbf D_2 \otimes \mathbf I_{n_{p_1}}$, we can write in the model's directional difference coefficient form, \begin{equation}
    (\mathbf{z}_y)_p = (\mathbf{D}_y\mathbf{m}')_p,
    \qquad
    (\mathbf{z}_x)_p = (\mathbf{D}_x\mathbf{m}')_p .
    \label{eq:coeffs_are_diffs}
\end{equation}
In synthesis form, using the fact that $(\mathbf A \otimes  \mathbf B)^{-1} = \mathbf A^{-1} \otimes  \mathbf B^{-1}$, and by analogy with the 1D case, this suggests a dictionary of two separate directional integrators
\begin{equation}
	    \mathbf{G}_y = \mathbf{I}_{n_{p_2}}\otimes\mathbf{\Sigma}_1,
    \qquad
	 \mathbf{G}_x = \mathbf{\Sigma}_2\otimes\mathbf{I}_{n_{p_1}},
    \label{eq:directional_atoms}
\end{equation}
where $\mathbf{\Sigma}_1$ and $\mathbf{\Sigma}_2$ are the one-dimensional cumulative-sum integrators of the form~\eqref{eq:C_operator} acting along each axis. Explicitly, this implies that the model $\mathbf m$ is the superposition of cumulative sum basis functions (which are discrete step functions in the column space of~\eqref{eq:C_operator}), in the $y$ and $x$ directions. The coefficients $\mathbf z$ are the weights on these basis functions. Such integration atoms require care over a single degree of freedom: the overall level of the model. In this case, we set to zero each directional integrator's constant of integration -- i.e., we set $\frac{1}{\eta}$ to 0 in~\eqref{eq:xprime_Cz} where the integrator is given by \eqref{eq:C_operator}, since its inverse is not required. We then use one shared level by appending one constant atom -- the all-ones column $\mathbf{1}$, and free only this column from the lasso penalty. The minimal directional-difference dictionary in $d$ dimensions therefore comprises the $d$ directional integrators together with one shared free level, $K=d+1$ atoms. For two-dimensional images,
\begin{equation}
    \mathbf{C}
    =
    \begin{bmatrix}
        \mathbf{G}_y & \mathbf{G}_x & \mathbf{1}
    \end{bmatrix},
    \qquad
    \bm\Lambda=\operatorname{diag}(\mathbf{1},\mathbf{1},0),
    \label{eq:dictionary_2D_TV}
\end{equation}
so that $K=3$, the penalty weight freeing only the single column $\mathbf{1}$. Orthonormal wavelet dictionaries on the other hand, do not require a separate constant atom, since such scaling or shifting coefficients are already present in the coarse level of the model. This level is freed by setting the corresponding penalty weight to zero within the single wavelet atom ($K=1$, if only one family is used), as proposed in Section~\ref{sec:weighting}. The step from $K=1$, the separable invertible operator, to $K\geq 3$ is our progression from mixed, ``TV-like'' regularisation to genuine stacked anisotropic total variation \citep{esedoglu2004decomposition}. It is obtained here by moving \eqref{eq:standard_2D_TV} into an overcomplete synthesis dictionary.
\begin{figure}
	\centering
	\includegraphics[width=1\linewidth]{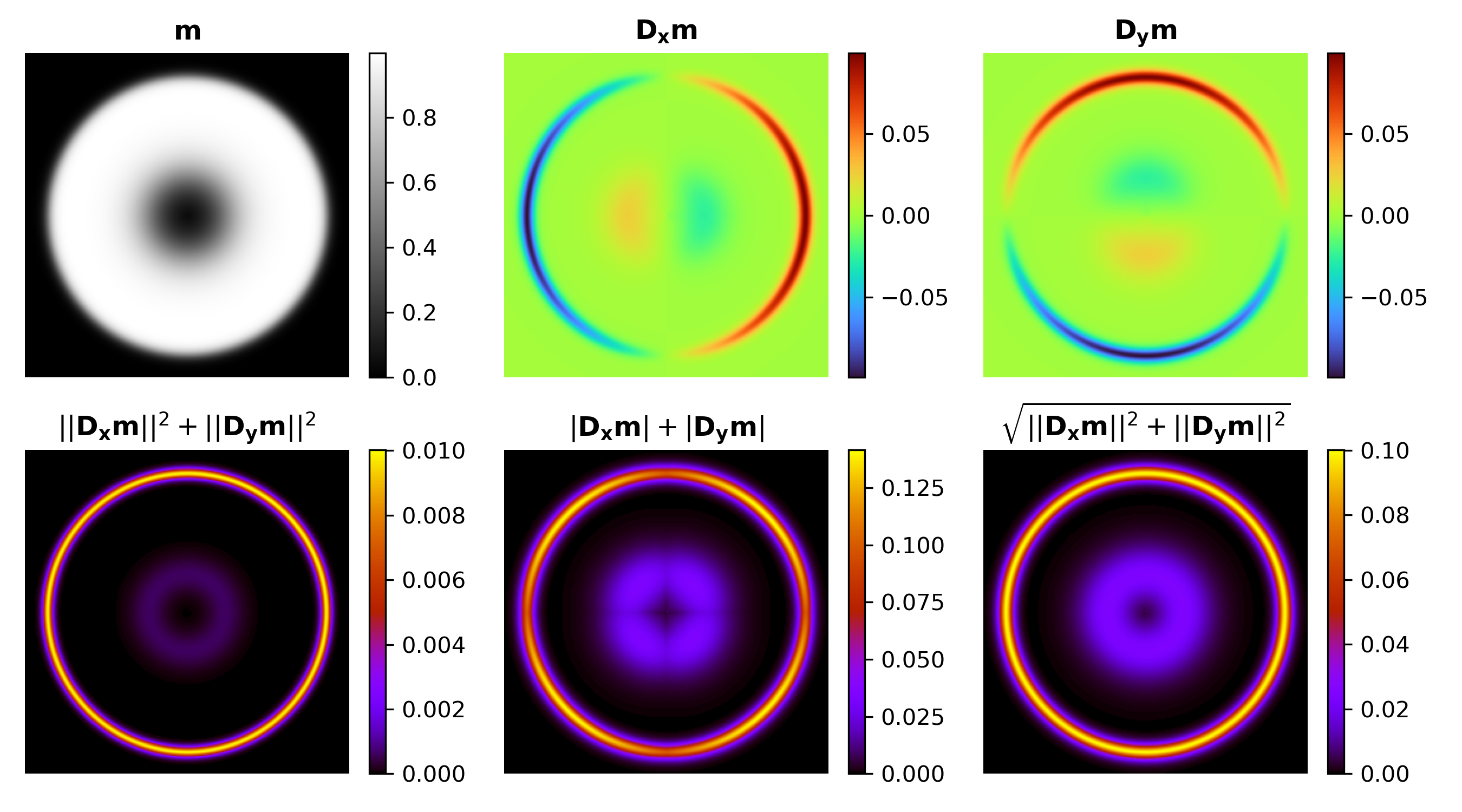}
	\caption{An image and various regularisation penalties. On the top row from left to right, are an image and its directional derivatives. Note that the image is an annulus with a blurred inside edge and a relatively sharp outside edge. The bottom row from left to right shows at every pixel, the ordinary $\ell_2$ regulariser, the anisotropic $\ell_1$ regulariser, and the isotropic $\ell_1$ regulariser. The $\ell_2$ and isotropic TV $\ell_1$ regularisers are not sensitive to edge orientation, while the anisotropic TV $\ell_1$ regulariser clearly has higher values for edges that are not axis aligned. Thus, anisotropic TV undesirably suppresses diagonal jumps. Note also in the bottom row, within each regulariser plot, the relative highlighting of edges for the inner and outer part of the annulus. $\ell_2$ penalises the sharp outer edge more strongly than its square-root-per-pixel counterpart (the $\ell_1$ isotropic TV regulariser) does. There is a factor of $\sqrt 2$ between the maximums of anisotropic and isotropic TV, as discussed in the text.}
	\label{fig:regdiff}
\end{figure}
Since the directional atoms synthesise by integration, we can write~\eqref{eq:standard_2D_TV} as a penalty on the model's spatial gradients. From~\eqref{eq:coeffs_are_diffs} we see that anisotropic TV penalises the two directional components of each pixel $p$ separately with an $\ell_1$ penalty,
\begin{equation}
    \mathrm{TV}_{\mathrm{aniso}}(\mathbf{z})
    =
    \sum_p \abs{(\mathbf{z}_y)_p} + \sum_p \abs{(\mathbf{z}_x)_p}.
    \label{eq:aniso}
\end{equation}
This is what the directional dictionary penalises directly through any lasso scheme. For a diagonal step, for a change $a$ in each dimension,~\eqref{eq:aniso} assigns a penalty $\abs{a}+\abs{a}=2a$. 
\subsection{Isotropic total variation}
Isotropic TV instead penalises the per-pixel gradient magnitude,
\begin{equation}
    \mathrm{TV}_{\mathrm{iso}}(\mathbf{z})
    =
    \sum_p \sqrt{(\mathbf{z}_y)_p^2 + (\mathbf{z}_x)_p^2}.
    \label{eq:iso}
\end{equation}
As shown in Figure~\ref{fig:regdiff}, the two directions enter through a single square root: making it $\ell_2$ within a pixel and $\ell_1$ across pixels. A diagonal step costs $\sqrt{a^2+a^2}=\sqrt2\,a$ proportional to the Euclidean diagonal distance, rather than $2a$, proportional to the $\ell_1$ diagonal distance. This ``over-penalisation'' of non-axis aligned features in anisotropic TV is undesirable in a Euclidean geometrical setting, which is why isotropic TV is preferred. Further, isotropic TV is approximately invariant to edge orientation, i.e., it works well unless a diagonal feature is very fine with each directional gradient discretised onto separate pixels. Since the square root couples the two components, isotropic TV is not a separable per-coefficient penalty solvable through the lasso. A reweighted-$\ell_2$ scheme (weighted least squares) converges to isotropic TV~\eqref{eq:iso} through a sequence of steps $k$. We will now consider the isotropic TV functional, and the use of a surrogate to approximate it.

\subsection{Reweighted least-squares surrogate}
\begin{figure}
	\centering
	\includegraphics[width=0.5\linewidth]{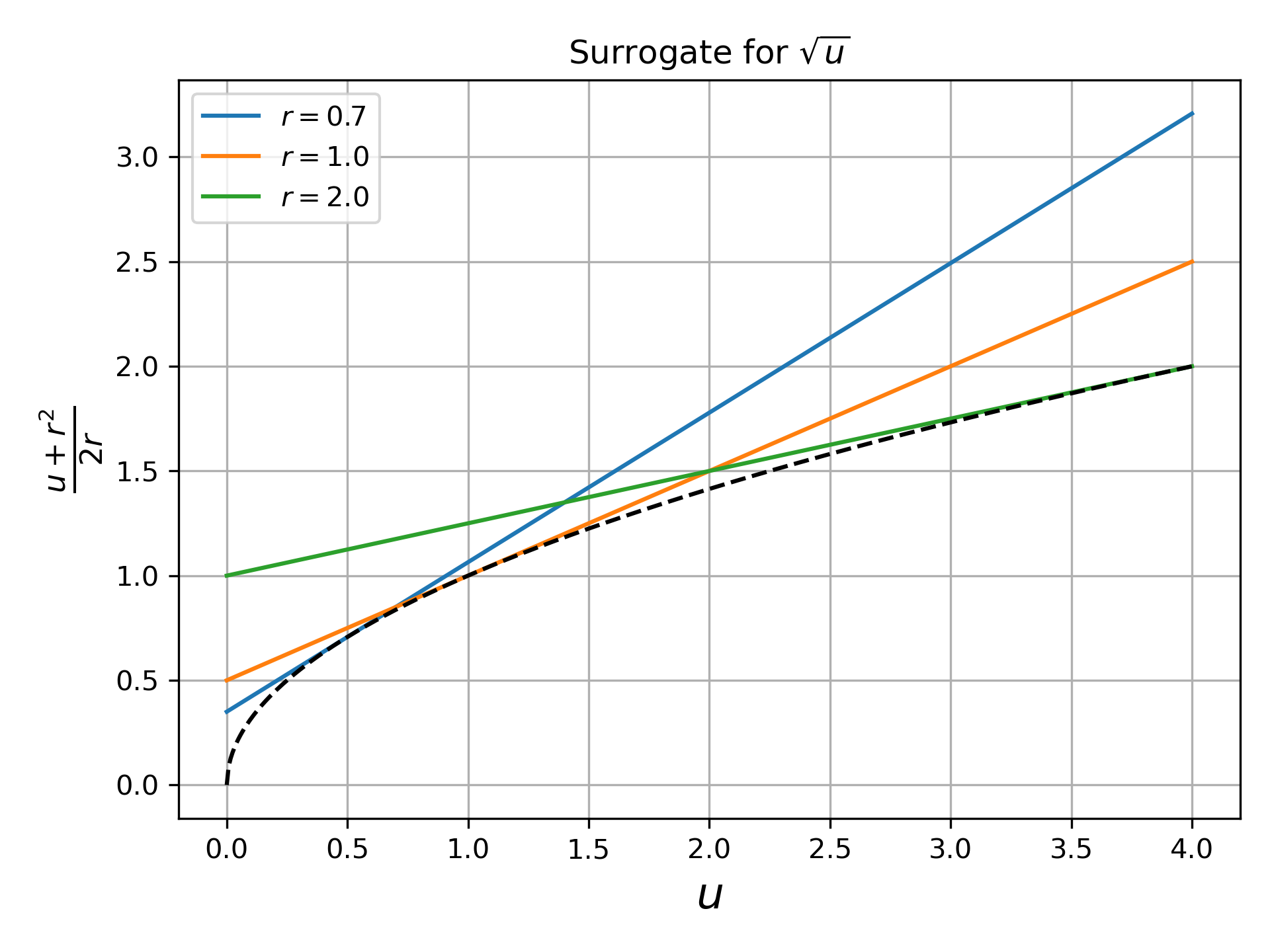}
	\caption{The upper bound or majorant that is used as a surrogate of $\sqrt u$ at distinct values of the parameter $r$. The surrogate only ever touches the square root of $u$ at $u=r^2$.}
	\label{fig:sqr_surr}
\end{figure}
The isotropic TV functional \eqref{eq:iso} is convex in $\mathbf z$, but the square root suffers from the same differentiability issues as the absolute value norm. This is where an easily differentiable surrogate function is useful. As shown in Figure~\ref{fig:sqr_surr}, the majorant of the square root is, for $u\ge0$ and any $r>0$ \citep{bach2012optimization},
\begin{equation}
    \sqrt{u} \;\le\; \frac{u + r^2}{2r},
    \qquad\text{with equality at } r=\sqrt{u}, \label{eqn:surrogate}
\end{equation}
so $\sqrt{u}$ is bounded above, i.e., majorised, by a function linear in $u$ that touches it at the current
point $r$. Applying this to~\eqref{eq:iso} per pixel with
\begin{equation}
u = (\mathbf{z}_y)_p^2 + (\mathbf{z}_x)_p^2,
\end{equation}
and the frozen current magnitude at step $k$ defined as
\begin{equation}
    r_p^{(k)}
    =
    \sqrt{(\mathbf{z}_y^{(k)})_p^2 + (\mathbf{z}_x^{(k)})_p^2 + \varepsilon},
    \qquad \varepsilon>0,
    \label{eq:rp}
\end{equation}
gives the reweighted-$\ell_2$ surrogate of the isotropic TV term (dropping
constants that do not depend on $\mathbf{z}$),
\begin{equation}
    \mathrm{TV}_{\mathrm{iso}}(\mathbf{z})
    \;\le\;
    \tfrac{1}{2}\sum_p \Omega_p^{(k)}
    \left[ (\mathbf{z}_y)_p^2 + (\mathbf{z}_x)_p^2 \right]
    + \text{const},
    \qquad
    \Omega_p^{(k)} = \frac{1}{r_p^{(k)}} .
    \label{eq:surrogate}
\end{equation}
The per-pixel weight $\Omega_p^{(k)}$ is computed directly from the current coefficients via~\eqref{eq:rp} and~\eqref{eq:coeffs_are_diffs}, and is applied to both directional coefficients of pixel $p$. A pixel whose gradient is large in either direction receives a small weight, relaxing the penalty on both of its components together. When the frozen weight is consistent with the coefficients it multiplies, 
\begin{equation}
r_p^{(k)}=\sqrt{(\mathbf{z}_y)_p^2+(\mathbf{z}_x)_p^2+\varepsilon}.
\end{equation}
Now adding back in the dropped $r_p/2$ constant term, for small $\varepsilon$, each pixel contributes in \eqref{eq:surrogate}:
\begin{equation}
    \frac{(\mathbf{z}_y)_p^2 + (\mathbf{z}_x)_p^2}
         {\sqrt{(\mathbf{z}_y)_p^2 + (\mathbf{z}_x)_p^2 + \varepsilon}}
    \;\approx\;
    \sqrt{(\mathbf{z}_y)_p^2 + (\mathbf{z}_x)_p^2},
\end{equation}
recovering the isotropic TV term~\eqref{eq:iso} up to the $\varepsilon$ smoothing. The iteration sequence -- compute $\Omega_p^{(k)}$ from $\mathbf{z}^{(k)}$, solve the weighted least-squares problem, then update $\mathbf{z}^{(k+1)}$, is a majorise-minimise scheme whose fixed point is the $\varepsilon$-smoothed isotropic TV solution \citep{hunter2004tutorial}. A small fixed $\varepsilon \approx 10^{-6}$ is used for stability, $\varepsilon\to0$ sharpens toward exact isotropic TV.

\subsection{The least squares subproblem}

With the weights frozen, the surrogate is quadratic in $\mathbf{z}$, so the regularised inverse problem at iterate $k$ is a damped weighted least-squares problem,
\begin{equation}
    \mathbf{z}^{(k+1)}
    =
    \argmin_{\mathbf{z}}
    \;
    \norm{\mathbf{y} - \mathbf{A}\mathbf{C}\mathbf{z}}^2
    +
    \lambda^2 \sum_p \Omega_p^{(k)}
    \left[(\mathbf{z}_y)_p^2 + (\mathbf{z}_x)_p^2\right],
    \label{eq:cgls_subproblem}
\end{equation}
At this juncture, we note that we do not directly use the synthesis framework in $\mathbf z$ for isotropic TV as there is no operator $\mathbf{C}$ for synthesis acting on coefficients $\mathbf z$. By~\eqref{eq:coeffs_are_diffs} the coefficient penalty 
\begin{equation}
\sum_p \Omega_p^{(k)}[(\mathbf{z}_y)_p^2+(\mathbf{z}_x)_p^2],
\end{equation}
equals the model-gradient penalty 
\begin{equation}
\sum_p \Omega_p^{(k)}[(\mathbf{D}_y\mathbf{m}')_p^2+(\mathbf{D}_x\mathbf{m}')_p^2]
= \norm{\sqrt{{\bm\Omega}_{r^{(k)}}}\,\mathbf{D}\mathbf{m}'}^2,
\end{equation}
with
$\mathbf{D}=[\mathbf{D}_y;\mathbf{D}_x]$ and $\bm\Omega_{r^{(k)}}=\mathrm{diag}(\Omega_p^{(k)})$
applied to both directional components of each pixel. Since we are not interested in the invertibility of $\mathbf D$ in this formulation, we can again safely set $\eta$ in \eqref{eq:D_operator} to 0. This is the ``lagged diffusivity'' form \citep{vogel1996iterative} of reweighted-$\ell_2$ isotropic TV: the weight $\Omega_p^{(k)}=1/r_p^{(k)}$ is small across edges (promotes little smoothing) and large in flat regions (leads to strong smoothing). This is yet another angle to view the reweighting sequence from: Weights are low in regions of high gradient from the previous weighted model. There is low impetus to the current reweighting iteration in~\eqref{eq:cgls_subproblem} to change pixels with already identified edges. Conversely, parts of the image with low gradient from the previous iteration, stand out for the regulariser in~\eqref{eq:cgls_subproblem} to smooth out. By the time the fixed point is reached, edges stand out sharply and the image is ``smooth'' in a total variation sense. For isotropic TV, we directly use in imaging parlance the ``analysis'' (as opposed to synthesis) form in \eqref{eq:fused_lasso_shifted} with $\beta^2 = 0$: 
\begin{equation}
    \min_{\mathbf{m}^{(k+1)}}
    \norm{
    \begin{bmatrix} \mathbf{A} \\[2pt]
                    \lambda\,\sqrt{\bm{\Omega}_{r^{(k)}}}\,\mathbf{D} \end{bmatrix}
    \mathbf{m}^{(k+1)}
    -
    \begin{bmatrix} \mathbf{y} \\[2pt] \mathbf{0} \end{bmatrix}
    }^2,
    \label{eq:cgls_stacked}
\end{equation}
solved by the least squares minimum residual \citep[LSMR,][]{fong2011lsmr} method without forming the normal equations \citep{paige1982lsqr}. The primes have been dropped in~\eqref{eq:cgls_stacked} as we do not incorporate a reference model for the $\ell_1$ fusion problem, setting $\beta^2=0$. The need for multiple steps $k$ in arriving at the isotropic TV solution is the following. The full TV objective function being the sum of a quadratic data misfit norm and the TV functional is convex, but not quadratic. The reweighted equivalent problems \eqref{eq:cgls_subproblem} or \eqref{eq:cgls_stacked} are quadratic. Therefore the reweighted least squares solutions must approach the TV solution through the majorisation-minimisation process. The only model-dependent ingredient is the diagonal weight $\bm{\Omega}_{r^{(k)}}$, which we recompute \textit{once} at every single $\lambda^2$ within a sweep, instead of calculating it multiple times within a separate inner iteration loop per $\lambda^2$. This is where we depart from standard IRLS, which saves us one inner loop, since the aim is not to converge to isotropic TV at every iteration of $k$, but by the end of the $\lambda^2$ sweep. We simply use a LSMR solver for the stacked system~\eqref{eq:cgls_stacked} while we sweep over $\lambda^2$ from high to low values as is usually done in Occam.
\section{An IRLS based interlude}
\begin{figure}
	\centering
	\includegraphics[width=0.5\linewidth]{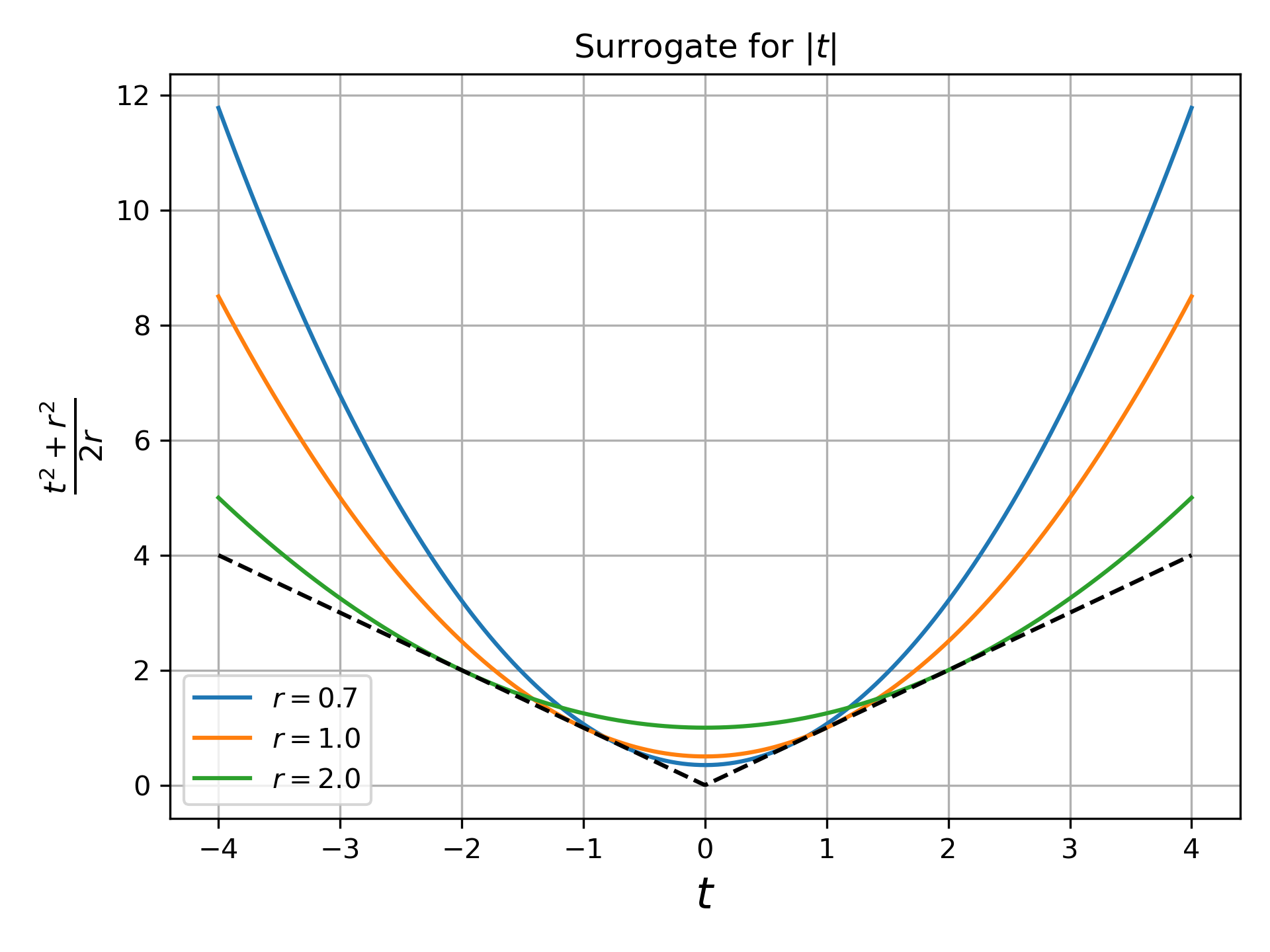}
	\caption{The majorant of $|t|$ at distinct values of the parameter $r$. In this case, the surrogate touches the absolute value of $t$ when $r=|t|$.}
	\label{fig:abs_surr}
\end{figure}
Re-examining \eqref{eqn:surrogate}, we can write it in a slightly different form \citep{bach2012optimization},
\begin{equation}
    |t| \;\le\; \frac{t^2 + r^2}{2r},
    \qquad\text{with equality at } r=|t|. \label{eqn:eta2}
\end{equation}
Now examining Figure~\ref{fig:abs_surr}, we realise that the absolute value function can also be approximated in the same way as the square root was for isotropic total variation. Using the surrogate~\eqref{eqn:eta2} for the lasso regularisation term $|\mathbf z | = \mathrm{L1_\text{lasso}}(\mathbf{z})$, with 
\begin{equation}
t = |\mathbf z_p|,
\end{equation}
per coefficient $p$ and defining the frozen current magnitude at step $k$ as
\begin{equation}
    r_p^{(k)}
    =
    \abs{\mathbf z_p^{(k)}} + \varepsilon,
    \qquad \varepsilon>0,
    \label{eq:rp_lasso}
\end{equation}
we arrive at the reweighted-$\ell_2$ surrogate of the $\ell_1$ term (dropping constants that do not depend on $\mathbf{z}$), 
\begin{equation}
    \mathrm{L1_\text{lasso}}(\mathbf{z})
    \;\le\;
    \tfrac{1}{2}\sum_p \Omega_p^{(k)}
    \mathbf{z}_p^2 
    + \text{const},
    \qquad
    \Omega_p^{(k)} = \frac{1}{r_p^{(k)}} .
    \label{eq:surrogate_L1}
\end{equation}
When the frozen weight is consistent with the coefficients it multiplies, $r_p^{(k)}=\abs{\mathbf{z}_p} + \varepsilon$. Again, on substituting this consistent weight into \eqref{eq:surrogate_L1} and adding back the $r_p/2$ constant, each pixel contributes:
\begin{equation}
    \frac{\mathbf{z}_p^2}
         {\abs{\mathbf{z}_p} + \varepsilon}
    \;\approx\;
    \abs{\mathbf{z}_p},
\end{equation}
recovering the $\ell_1$ regularisation term up to the $\varepsilon$ smoothing \citep{candes2008enhancing}. As in the isotropic TV case, the iteration sequence: compute $\Omega_p^{(k)}$ from $\mathbf{z}^{(k)}$, solve the weighted least-squares problem, finally update $\mathbf{z}^{(k+1)}$, is again a majorise--minimise scheme whose fixed point is the $\varepsilon$-smoothed $\ell_1$ lasso solution \citep{daubechies2010iteratively}. A small fixed $\varepsilon\approx10^{-6}$ is used for stability, $\varepsilon\to0$ sharpens toward exact $\ell_1$ Lasso.
\subsection{The least squares subproblem for lasso}
We can now write \eqref{eq:general_D_lasso} or \eqref{eq:synthesis_lasso_problem} as 
\begin{equation}
    \mathbf{z}^{(k+1)}
    =
    \argmin_{\mathbf{z}}
    \;
    \norm{\mathbf{y} - \mathbf{T}\mathbf{z}}^2
    +
    \lambda^2 \sum_p \Omega_p^{(k)}
    \mathbf{z}_p^2,
    \label{eq:cgls_subproblem_lasso}
\end{equation}
and this form works for both the invertible and synthesis cases. With $\mathbf{T=AC}$  we solve
\begin{equation}
    \min_{\mathbf{z}}
    \norm{
    \begin{bmatrix} \mathbf{AC} \\[2pt]
                    \lambda\,\sqrt{\bm{\Lambda}_{r^{(k)}}}\end{bmatrix}
    \mathbf{z}
    -
    \begin{bmatrix} \mathbf{y} \\[2pt] \mathbf{0} \end{bmatrix}
    }^2,
    \label{eq:cgls_stacked_lasso}
\end{equation}
again using LSMR without forming the normal equations. The only model-dependent ingredient is the diagonal weight $\bm{\Lambda}_{r^{(k)}} = \mathrm{diag}\left(\Omega_p^{(k)}\Lambda_p \right)= \mathrm{diag}\left(\frac{\Lambda_p}{r_p^{(k)}}\right)$, where $\Lambda_p$ without a superscript $^{(k)}$ indicates the unchanging prior weight (Section~\ref{sec:weighting}) for the lasso problem on coefficient $(\mathbf z)_p$. We recompute the weights $\frac 1 {r_p^{(k)}}$ once at every single $\lambda^2$ within a sweep, as we did for the isotropic TV case. 

\section{Split Bregman methods for $\ell_1$ regularisation}
\label{sec:bregman}

The reweighted-$\ell_2$ schemes behind the surrogates~\eqref{eq:surrogate} and~\eqref{eq:surrogate_L1} get around the non-differentiability of the $\ell_1$ term by replacing it with a smooth quadratic surrogate. This is done through majorisation-minimisation, with all the machinery using conventional $\ell_2$ least squares solves. Split Bregman instead splits the problem into its $\ell_2$ and $\ell_1$ parts, solved in alternate steps, followed by an update which cascades through both parts \citep{goldstein2009split}. It is equivalent to the alternating-direction method of multipliers, or a Bregman or
augmented-Lagrangian iteration \citep{boyd2011distributed} as explained underneath. We consider the problem of the form \eqref{eq:z_objective_full_weighted} with $\beta^2=0$ but without requiring invertibility of $\mathbf{D}$,
\begin{equation}
    \argmin_{\mathbf m}\ \phi(\mathbf {m}) = 
    \frac 1 2\norm{\mathbf y - \mathbf{A}\mathbf m}^2
    + \lambda^2\,\norm{\mathbf {\bm\Lambda Dm}}_1.
    \label{eq:sb_tv}
\end{equation}
Let us include a quadratic penalty that enforces the $\ell_1$ term exactly as $\gamma\rightarrow\infty$ through an auxiliary variable $\mathbf d$
\begin{equation}
    \min_{\mathbf m,\mathbf d}  \phi(\mathbf {m, d}) = 
    \frac 1 2 \norm{\mathbf y - \mathbf{A}\mathbf m}^2
    + \lambda^2\,\norm{\bm\Lambda \mathbf d}_1
    + \frac{\gamma}{2}\norm{\mathbf{Dm-d}}^2.
    \label{eq:sb_nolag}
\end{equation}
Unfortunately, large values of $\gamma$ can lead to conditioning problems in \eqref{eq:sb_nolag} especially if $\mathbf D$ contains difference operators with a nullspace. We consider instead the augmented Lagrange problem with multiplier $\bm \mu$,
\begin{equation}
    \min_{\mathbf m,\mathbf d} \phi(\mathbf {m, d}) = 
    \frac 1 2 \norm{\mathbf y - \mathbf{A}\mathbf m}^2
    + \lambda^2\,\norm{\bm\Lambda \mathbf d}_1
    + \bm\mu^t(\mathbf{Dm-d})
    + \frac{\gamma}{2}\norm{\mathbf{Dm-d}}^2,
    \label{eq:sb_auglag}
\end{equation}
where $\bm\mu$ is the Lagrange multiplier enforcing $\mathbf{Dm=d}$, thus making $\frac{\gamma}{2}\norm{\mathbf{Dm-d}}^2$ the augmentation. Writing $\mathbf r = \mathbf{Dm-d}$ and completing squares on the resulting $\bm\mu ^t \mathbf r + \frac \gamma 2 \mathbf {r}^t\mathbf{r}$ terms in \eqref{eq:sb_auglag}, dropping constant terms we arrive at
\begin{equation}
    \min_{\mathbf m,\mathbf d} \phi(\mathbf {m, d}) = 
    \frac 1 2 \norm{\mathbf y - \mathbf{A}\mathbf m}^2
    + \lambda^2\,\norm{\bm\Lambda \mathbf d}_1
    + \frac{\gamma}{2}\norm{\mathbf{Dm-d}+\frac{\bm\mu} \gamma }^2,
    \label{eq:sb_auglag_no_b}
\end{equation}
now defining the Bregman variable
\begin{equation}
\mathbf{b} = \frac{\bm\mu} \gamma, \label{eq:bregman_b}
\end{equation}
we arrive at the more familiar split Bregman form:
\begin{equation}
    \min_{\mathbf m,\mathbf d} \phi(\mathbf {m, d}) = 
    \frac 1 2 \norm{\mathbf y - \mathbf{A}\mathbf m}^2
    + \lambda^2\,\norm{\bm\Lambda \mathbf d}_1
    + \frac{\gamma}{2}\norm{\mathbf{Dm-d}+\mathbf b }^2.
    \label{eq:sb}
\end{equation}
As proposed by \cite{goldstein2009split}, splitting the problem into the $\ell_2$ and $\ell_1$ parts, if we take the gradient with respect to $\mathbf m$, we find for optimality, that
\begin{equation}
	\mathbf{\left(A^\tran A + \gamma D^\tran D\right)\hat m} = \mathbf{A^\tran y + \gamma D^\tran (d-b)}. \label{eq:sb1}
\end{equation}
Using the value of $\mathbf m$ thus obtained, for the $\ell_1$ part, writing
\begin{equation}
	\mathbf{v=D\hat m+b},
\end{equation}
we recast \eqref{eq:sb} as
\begin{equation}
	\min_{\mathbf d} \frac \gamma 2 \norm{\mathbf {v-d}}^2 + \lambda^2\norm{\bm\Lambda\mathbf d}_1 \label{eq:sb_l1}
\end{equation}
the solution of which, as shown in \eqref{eq:lasso_componentwise_independent} of Appendix~\ref{sec:whycd}, after accommodating the factor of $\gamma$ on the data norm in \eqref{eq:sb_l1} is simply 
\begin{equation}
	\hat {\mathbf d}_p = S\left(v_p, \frac{\lambda^2 \Lambda_p} \gamma\right), \qquad \forall p = 1, \dots, n_p, \label{eq:sb2}
\end{equation}
where $S$ is the soft threshold function \eqref{eq:soft_threshold_scalar}, and $\Lambda_p$ are static weights. We note that in \eqref{eq:sb2}, no costly coordinate descent is required, as each component $p$ can be soft thresholded independent of any other component. 

This brings us to the last step of the split Bregman method, the update step. From \eqref{eq:sb_auglag}, we can see that the stationary point with respect to $\mathbf{m}$ for both the augmented and unaugmented form should be the same. In the absence of the augmentation $\frac{\gamma}{2}\norm{\mathbf{Dm-d}}^2$, the stationary point in $\mathbf m$ satisfies
\begin{equation}
	\mathbf{A^\tran\left(Am-y\right)} + \mathbf{D^\tran \bm\mu} = \mathbf 0, \label{eq:stationarity_unaugmented}
\end{equation}
and we know from \eqref{eq:sb1} and \eqref{eq:bregman_b} that the augmented form satisfies 
\begin{equation}
	\mathbf{A^\tran\left(Am-y\right)} + \mathbf{D^\tran\left(\gamma b + \gamma (Dm - d)\right)} = \mathbf 0, \label{eq:stationarity_augmented}
\end{equation}
with the Lagrange multiplier having shifted. Using \eqref{eq:bregman_b} we can now write the correction from the above two equations as
\begin{equation}
	\mathbf b_\mathrm{new} = \frac {\bm \mu}{\gamma} = \mathbf{b + D\hat m - \hat d} . \label{eq:bregmanacc}
\end{equation}
For consistency, we can check that when the constraints $\mathbf{D m \approx d}$ are met, the fixed point of \eqref{eq:sb} is then the same as that of \eqref{eq:sb_nolag}, which solves the original problem \eqref{eq:sb_tv}. Conversely, another way to look at it is that $\mathbf b_\mathrm{new} = \mathbf b$ when the constraints are satisfied. To summarise, split Bregman consists of three steps carried out in $k$ iterations:

\noindent
\textit{(i) Least squares step.}
The $\ell_2$ solution \eqref{eq:sb1} is again given in analysis form by 
\begin{equation}
    \mathbf m^{(k+1)}
    =
    \argmin_{\mathbf m}\;
    \norm{
    \begin{bmatrix} \mathbf{A} \\[2pt] \sqrt{\gamma}\,\mathbf D \end{bmatrix}
    \mathbf m
    -
    \begin{bmatrix} \mathbf y \\[2pt] \sqrt{\gamma}\,(\mathbf d^{(k)} - \mathbf b^{(k)}) \end{bmatrix}
    }^2,
    \label{eq:sb_stacked}
\end{equation}
solved by LSMR without forming the normal equations, exactly as for the reweighted stack \eqref{eq:cgls_stacked}.

\noindent
\textit{(ii) Uncoupled lasso step.} With $\mathbf m$ frozen the $\mathbf d$-subproblem separates per coefficient and is solved in \eqref{eq:sb2} by soft-thresholding,
\begin{equation}
    \mathbf d_p^{(k+1)}
    = S\!\left( (\mathbf {Dm}^{(k+1)} + \mathbf b^{(k)})_p,\; \tau_p \right),
    \qquad
    \tau_p = \frac{\lambda^2 \Lambda_p}{\gamma},
    \label{eq:sb_shrink}
\end{equation}
The static prior $\Lambda_p$ enters here, as a per-coefficient threshold, rather than inside the operator.

\noindent
\textit{(iii) Bregman update.}
\begin{equation}
    \mathbf b^{(k+1)} = \mathbf b^{(k)} + \left( \mathbf {Dm}^{(k+1)} - \mathbf d^{(k+1)} \right).
    \label{eq:sb_update}
\end{equation}
Steps (i)--(iii) are iterated to convergence, for specific forms as discussed next.
\subsection{Invertible and synthesis forms}
When $\mathbf{D}$ is invertible, as earlier, a change of variables $\mathbf{z}=\mathbf{D}\mathbf{m}$, $\mathbf{C}=\mathbf{D}^{-1}$ converts \eqref{eq:sb_tv} into the lasso \eqref{eq:general_D_lasso}, with the model recovered through the synthesis map $\mathbf{m}=\mathbf{C}\mathbf{z}$. When $\mathbf{D}$ is not invertible, we have a synthesis dictionary as before, with $ \mathbf{m}
    =
    \mathbf{C}\mathbf{z}
    =
    \sum_{i=1}^{K}
    \mathbf{G}_i\mathbf{z}_i
    $. We then return to the lasso problem \eqref{eq:synthesis_lasso_problem} through the application of \eqref{eq:synthesis_T}, by regularising $\mathbf z$ in place of $\mathbf{Dm}$ in~\eqref{eq:sb_tv} and~\eqref{eq:sb_nolag}. In both cases we proceed with Bregman iterations exactly as we have done in~\eqref{eq:sb_stacked}-\eqref{eq:sb_update}.
\subsection{Anisotropic total variation}
\label{sec:sb_aniso}
With Bregman variables $\mathbf b_i$ corresponding to the directional gradients $\mathbf d_i = \mathbf {D}_i \mathbf {m}$ for $i\in\{y,x\}$, the least-squares step has a stacked operator $\mathbf D$ in  \eqref{eq:sb_stacked}. Writing
\begin{equation}
	v_{i,p} = (\mathbf D_i\mathbf m^{(k+1)} + \mathbf b_i^{(k)})_p,
\end{equation}
the per-pixel shrinkage step can explicitly be written, per component $i$, as:
\begin{equation}
    s_{i,p} = |v_{i,p}|,
    \qquad
     d_{i,p}^{(k+1)}
    = \max\!\left( s_{i,p} - \tfrac{\lambda^2\Lambda_p}{\gamma},\; 0 \right)
      \frac{v_{i,p}}{s_{i,p}},
    \label{eq:sb_aniso_shrink}
\end{equation}
and this produces the anisotropic TV solution.
\subsection{Isotropic total variation by splitting}
\label{sec:sb_iso}
If instead we couple the directional gradients through their $\ell_2$ magnitude per pixel, we get,
\begin{equation}
    s_p = \sqrt{\sum_i v_{i,p}^2},
    \qquad
    d_{i,p}^{(k+1)}
    = \max\!\left( s_p - \tfrac{\lambda^2\Lambda_p}{\gamma},\; 0 \right)
      \frac{v_{i,p}}{s_p},
    \label{eq:sb_iso_shrink}
\end{equation}
we arrive at the isotropic TV solution. \eqref{eq:sb_iso_shrink} can be thought of as the magnitude of the pixel gradients, times the unit vector for the component per pixel.

In both TV cases, just as in IRLS, $\eta$ can be set to zero as we are not interested in the invertibility of $\mathbf{D}=[\mathbf{D}_y;\mathbf{D}_x]$. The soft thresholding step is followed by the Bregman updates $\mathbf b_i^{(k+1)} = \mathbf b_i^{(k)} + (\mathbf D_i\mathbf m^{(k+1)} - \mathbf d_i^{(k+1)})$. Note how it is only a change in the shrinkage norm per pixel from $\norm{}_1$ for $s_{i,p}$, to $\norm{}_2$ for $s_{p}$ that produces anisotropic or isotropic TV, respectively \citep{wang2007caam, wang2008new}. The sum over $i$ in~\eqref{eq:sb_aniso_shrink} and \eqref{eq:sb_iso_shrink} extends unchanged to three dimensions.

\subsection{Relation to IRLS}
\label{sec:sb_relation}
Comparing the split Bregman stacked form \eqref{eq:sb_stacked} with the reweighted stacked forms \eqref{eq:cgls_stacked_lasso} and~\eqref{eq:cgls_stacked} exposes a duality of sorts. The forward part ($\mathbf{AC}$ or $\mathbf A$) of the operator block is fixed for both. The reweighted scheme then carries the $k$ iterate dependence in the operator block: the regularisation part $\lambda\sqrt{\bm\Lambda_r^{(k)}}$ of the operator block is rebuilt every step while the corresponding right-hand side part stays $\mathbf 0$. Split Bregman instead carries the iterate dependence, largely in the right-hand side $\sqrt{\gamma}(\mathbf d^{(k)}-\mathbf b^{(k)})$. We say largely because $\gamma \mathbf D$ in the operator for split Bregman will scale with a changing $\gamma$, but this term will not change the operator's column space as such since $\mathbf D$ itself does not change. Therefore, IRLS and split Bregman iterate differently: the former reweights the operator with a fixed right-hand side, while the latter leaves the operator structure largely unchanged and modifies the right-hand side.

The soft thresholding in \eqref{eq:sb_shrink} and~\eqref{eq:sb_iso_shrink} returns exact zeros, so split Bregman targets the exact $\ell_1$ or isotropic-TV functional with no smoothing parameter, whereas the reweighted IRLS surrogate reaches the $\varepsilon$-smoothed one of \eqref{eq:surrogate}. The penalty parameter $\gamma$ controls only the convergence rate, not the solution: any $\gamma>0$ yields the same minimiser of~\eqref{eq:sb_tv} at the fixed point. To simplify the $\tau_p$ threshold to 1 for iso-TV in \eqref{eq:sb_shrink}, we set $\gamma$ to equal $\lambda^2$, which is analogous to the choice made by \cite{vargas2025rambo}. As we will see in the Applications (Section~\ref{sec:apps}) this parameterisation has consequences. The choice made here is different from the empirical suggestion made by \cite{goldstein2009split}, since they weight the data misfit norm, not the $\ell_1$ TV regularisation term, and their recommendation would amount to a constant $\gamma$ in~\eqref{eq:sb} -- which our tests have found hard to tune. Since the Bregman iterations in \eqref{eq:sb_stacked} through \eqref{eq:sb_update} use a linked value of $\lambda^2$ and $\gamma$ it is necessary to reset $\mathbf b$ and $\mathbf d$ to zero for every new value of $\lambda^2$ being considered.

Following the alternating direction method of multipliers \citep[ADMM,][]{boyd2011distributed} convention, defining the primal residual as 
$
\Delta \mathbf b = \mathbf {D m}^{(k+1)} - \mathbf d^{(k+1)} 
$
it is easy to see that $\norm{\Delta \mathbf b} \rightarrow 0$ would imply that the constraints are being met in \eqref{eq:sb}. With a little algebra, it can be shown from \eqref{eq:stationarity_unaugmented} and \eqref{eq:stationarity_augmented}, that for both forms to be representative of the same stationary point in $\mathbf m$ after the Bregman update, the quantity  
$
\mathbf D^{\tran}\Delta \mathbf d,
$
must also go to zero. Even without using $\mathbf D^\tran$ we have found the following simple joint convergence criterion to be effective: $\max(\frac{\norm{\Delta \mathbf b}}{\norm{\mathbf b}}, \frac{\norm{\Delta \mathbf d}}{\norm{\mathbf d}}) < \mathrm{tol}$. Unlike the IRLS scheme, our attempts to interleave the $\lambda^2$ sweep with the inner iteration~\eqref{eq:sb_stacked}--\eqref{eq:sb_update} rather than run to convergence at each $\lambda^2$ met limited success, as discussed in the Applications section (Section~\ref{sec:apps}). 
\section{Nonlinear inverse problem}
\label{sec:nonlinear}

Finally, now consider the nonlinear inverse problem
\begin{equation}
    \mathbf{d} = \mathbf{f}(\mathbf{m}),
    \label{eq:nonlinear_forward}
\end{equation}
where $\mathbf{d}\in\mathbb{R}^{n}$ is the observed data vector, $\mathbf{m}\in\mathbb{R}^{p}$ is the model vector, and $\mathbf{f}(\cdot)$ is the nonlinear forward operator. Let $\mathbf{m}_1$ denote the current model and let $\mathbf{m}_2$ denote a candidate updated model. The first-order Taylor expansion of $\mathbf{f}$ about $\mathbf{m}_1$ gives
\begin{equation}
    \mathbf{f}(\mathbf{m}_2)
    \approx
    \mathbf{f}(\mathbf{m}_1)
    +
    \mathbf{J}
    \left(
        \mathbf{m}_2-\mathbf{m}_1
    \right),
    \label{eq:nonlinear_taylor_m2}
\end{equation}
where the Jacobian matrix of partial derivatives $\mathbf J \in\mathbb{R}^{n\times p}$ has elements
\begin{equation}
    \mathbf{J}_{ij}
    =
    \frac{\partial f_i}{\partial m_j}\Bigg|_{\mathbf m_1}.
    \label{eq:jacobian_definition}
\end{equation}
The nonlinear regularised objective for the candidate model $\mathbf{m}_2$ is
\begin{equation}
    \phi(\mathbf{m}_2)
    =
    \frac{1}{2}
    \left\|
        \mathbf{W}
        \left[
            \mathbf{f}(\mathbf{m}_2)-\mathbf{d}
        \right]
    \right\|_2^2
    +
    \lambda^2
    \left[
        \beta^2
        \left\|
            \mathbf{m}_2-\mathbf{m}_0
        \right\|_1
        +
        \left\|
            \mathbf{D}
            \left(
                \mathbf{m}_2-\mathbf{m}_0
            \right)
        \right\|_1
    \right],
    \label{eq:nonlinear_objective_m2}
\end{equation}
where $\mathbf{m}_0$ is the reference model. The data-weighting matrix $\mathbf{W}$ is related to the data covariance matrix $\mathbf{C}_d$ by
\begin{equation}
    \mathbf{W}^{T}\mathbf{W}
    =
    \mathbf{C}_d^{-1}.
    \label{eq:data_weight_covariance}
\end{equation}
To connect this nonlinear objective with the linear formulation discussed so far, we define
\begin{equation}
    \mathbf{f}_1 = \mathbf{f}(\mathbf{m}_1),
    \qquad
    \mathbf{m}' = \mathbf{m}_2-\mathbf{m}_0.
    \label{eq:nonlinear_shifted_model}
\end{equation}
Then
\begin{equation}
    \mathbf{m}_2 = \mathbf{m}'+\mathbf{m}_0.
    \label{eq:m2_from_mprime}
\end{equation}
Substituting equation~\eqref{eq:m2_from_mprime} into the linearisation
\eqref{eq:nonlinear_taylor_m2} gives
\begin{equation}
    \mathbf{f}(\mathbf{m}_2)
    \approx
    \mathbf{f}_1
    +
    \mathbf{J}
    \left(
        \mathbf{m}' + \mathbf{m}_0 - \mathbf{m}_1
    \right).
    \label{eq:linearised_forward_mprime}
\end{equation}
Thus
\begin{equation}
    \mathbf{f}(\mathbf{m}_2)-\mathbf{d}
    \approx
    \mathbf{J}\mathbf{m}'
    -
    \mathbf{r},
    \label{eq:linearised_residual_mprime}
\end{equation}
where
\begin{equation}
    \mathbf{r}
    =
    \mathbf{d}
    -
    \mathbf{f}_1
    +
    \mathbf{J}\mathbf{m}_1
    -
    \mathbf{J}\mathbf{m}_0.
    \label{eq:r_definition_nonlinear}
\end{equation}
The linearised objective \eqref{eq:nonlinear_objective_m2} may therefore be written as
\begin{equation}
    \phi(\mathbf{m}')
    \approx
    \frac{1}{2}
    \left\|
        \mathbf{W}
        \left(
            \mathbf{J}\mathbf{m}'-\mathbf{r}
        \right)
    \right\|_2^2
    +
    \lambda^2
    \left[
        \beta^2
        \left\|
            \mathbf{m}'
        \right\|_1
        +
        \left\|
            \mathbf{D}\mathbf{m}'
        \right\|_1
    \right].
    \label{eq:linearised_objective_mprime}
\end{equation}
Equivalently,
\begin{equation}
    \phi(\mathbf{m}')
    \approx
    \frac{1}{2}
    \left\|
        \mathbf{W}\mathbf{r}
        -
        \mathbf{W}\mathbf{J}\mathbf{m}'
    \right\|_2^2
    +
    \lambda^2
    \left[
        \beta^2
        \left\|
            \mathbf{m}'
        \right\|_1
        +
        \left\|
            \mathbf{D}\mathbf{m}'
        \right\|_1
    \right].
    \label{eq:linearised_objective_linear_form}
\end{equation}
This has the same form as the linear objective in \eqref{eq:fused_lasso_shifted}, with $\mathbf{y}'$, and $\mathbf{A}$ replaced by $\mathbf{W}\mathbf{r}$, and $\mathbf{W}\mathbf{J}$, respectively. Thus, each nonlinear iteration requires solving a linear fused-lasso problem of the same form considered above, and either the invertible, analysis, or synthesis parameterisation may be used to do so. At this juncture we note that $\mathbf W$ was kept out of the linear exposition to align with the lasso literature, and for notational simplicity. During implementation, both $\mathbf y$ and $\mathbf A$ were premultiplied by $\mathbf W$ even for the linear form \eqref{eq:fused_lasso_shifted} to ensure consistency between the linear and linearised problems.
\subsection{Analysis case}
The isotropic TV problem \eqref{eq:cgls_stacked} and \eqref{eq:sb_tv} after linearisation belong to the family described by \eqref{eq:linearised_objective_linear_form}, so no further transforms are necessary in the linear substep.  
\subsection{Invertible and synthesis cases}
For both the invertible and synthesis cases, the same linearisations \eqref{eq:linearised_forward_mprime}-\eqref{eq:r_definition_nonlinear} hold. Therefore, for the candidate model $\mathbf m_2$ we have
\begin{equation}
     \mathbf m'  = \mathbf{m}_2 - \mathbf{m}_0.
    \label{eq:synthesis_nl_m2}
\end{equation}
For the synthesis case, instead of reparameterising through an invertible $\mathbf{D}$, we now expand the perturbation in the synthesis dictionary \eqref{eq:synthesis_dictionary},
\begin{equation}
    \mathbf{m}'
    =
    \mathbf{C}\mathbf{z}
    =
    \sum_{i=1}^{K}
    \mathbf{G}_i\mathbf{z}_i ,
    \label{eq:synthesis_nl_model}
\end{equation}
and proceed with a unified treatment of both cases next. We write the deviation from the reference as $\mathbf{m}'=\mathbf{C}\mathbf{z}$, with
\begin{equation}
 \mathbf{C}=
\begin{cases}
\mathbf{D}^{-1}, & \text{invertible mode, } \mathbf{z}=\mathbf{D}\mathbf{m}', \\[1.0ex]
\begin{bmatrix}
        \mathbf{G}_1
        &
        \mathbf{G}_2
        &
        \cdots
        &
        \mathbf{G}_K
    \end{bmatrix}, & \text{synthesis mode}.
\end{cases}
    \label{eq:nonlinear_C_cases}
\end{equation}
Substituting the appropriate $\mathbf{m}'=\mathbf{C}\mathbf{z}$ into \eqref{eq:linearised_objective_linear_form}, recognising that the linearisation is the same for invertible and synthesis cases, using \eqref{eq:nonlinear_C_cases} we get the same objective function for both cases:
\begin{equation}
    \phi(\mathbf{z})
    \approx
    \frac{1}{2}
    \left\|
        \mathbf{W}\mathbf{r}
        -
        \mathbf{W}\mathbf{J}\mathbf{C}\mathbf{z}
    \right\|_2^2
    +
    \lambda^2
    \left[
        \beta^2
        \left\|
            \mathbf{C}\mathbf{z}
        \right\|_1
        +
        \left\|
            \bm\Lambda\mathbf{z}
        \right\|_1
    \right].
    \label{eq:nonlinear_z_objective_full}
\end{equation}
In \eqref{eq:nonlinear_z_objective_full} $\bm\Lambda$ is the usual diagonal penalty weight of Section~\ref{sec:weighting}. Defining the linearised forward matrix
\begin{equation}
    \mathbf{T}
    =
    \mathbf{W}\mathbf{J}\mathbf{C},
    \label{eq:nonlinear_T1_definition}
\end{equation}
the case $\beta^2=0$ becomes the weighted lasso problem
\begin{equation}
    \hat{\mathbf{z}}
    =
    \arg\min_{\mathbf{z}}
    \left\{
        \frac{1}{2}
        \left\|
            \mathbf{Wr}
            -
            \mathbf{T}\mathbf{z}
        \right\|_2^2
        +
        \lambda^2
        \left\|
            \bm\Lambda\mathbf{z}
        \right\|_1
    \right\},
    \label{eq:nonlinear_z_lasso_problem}
\end{equation}
of the same form \eqref{eq:general_D_lasso} as in the linear case, with the place of $\mathbf y'$ being taken by $\mathbf{Wr}$ in the linearisation above. The corresponding updated model is
\begin{equation}
    \hat{\mathbf{m}}_2
    =
    \mathbf{m}_0
    +
    \mathbf{C}\hat{\mathbf{z}}.
    \label{eq:nonlinear_m2_hat}
\end{equation}
For $\beta^2>0$ again, the soft-thresholded update is
\begin{equation}
    \mathbf{m}_{2,\mathrm{ST}}
    =
    \mathbf{m}_0
    +
    S
    \left(
        \mathbf{C}\hat{\mathbf{z}},
        \lambda^2\beta^2
    \right),
    \label{eq:nonlinear_m2_ST}
\end{equation}
where $S$ is the componentwise soft-thresholding operator defined in \eqref{eq:soft_threshold_scalar}.

\section{Occam phase two: selecting an optimal $\lambda^2$}
\label{sec:phase2}
In phase 2 of the Occam method, the $\chi^2$ misfit sampled along the high-to-low sweep over $\lambda^2$ crosses below the target value $\chi^2_*$. For the $\ell_2$ and coordinate descent $\ell_1$ routes, we use Brent's method and bracketing to find the $\lambda^2_*$ that corresponds to $\chi^2_*$. For the IRLS and split Bregman routes, enough reweighting or Bregman steps need to be run such that the weights or the auxiliary and Bregman variables stabilise through a sequence of secant steps (typically 5 or less). Therefore, we modify the original \cite{Constable1987} approach as follows: in accordance with the Occam interpretation of the Mozorov discrepancy principle \citep{morozov1968error}, the model with the highest $\lambda^2$ underneath the target $\chi^2_*$ is finally selected, if $\chi^2_*$ cannot be achieved within the maximum number of secant steps. For Gaussian data errors, $\chi^2_*$ is equal to the number of data, i.e., the expected $\chi^2_k$ value for $k$ observations. From a statistical point of view, any values close to this target are defensible, as the $\chi^2_k$ distribution has high probability in the vicinity of $k$. 
\section{Applications}
\label{sec:apps}
In this section, we will first consider a simple 1D linear subsampling and reconstruction problem, demonstrating the equivalence of various pieces of the mathematical machinery. While coordinate descent methods worked well for the 1D problem, they did not scale well to 2D because of matrix materialisation costs. We demonstrate scalability to higher dimensions with a blurred and subsampled image reconstruction problem using isotropic total variation and dictionary synthesis using the split Bregman and IRLS methods. Both split Bregman and IRLS avoid matrix materialisation through the use of LSMR. With a second image deblurring problem, we demonstrate how different prior assumptions for an inverse problem can lead to strikingly different results. Finally, we solve a nonlinear airborne electromagnetic (AEM) problem with first a synthetic sounding and then a flight line of field AEM data. To interpret the computational times reported, the synthetic examples were run single-threaded on an Apple M2 desktop @3.49 GHz. The field data were inverted in parallel on Intel Sapphire Rapids boards with each CPU running @1.9-2.2 GHz.

\subsection{A linear 1D problem: reconstruction from subsampled data}
\label{sec:1Dproblem}
\begin{figure}
	\centering
	\includegraphics[width=0.5\linewidth]{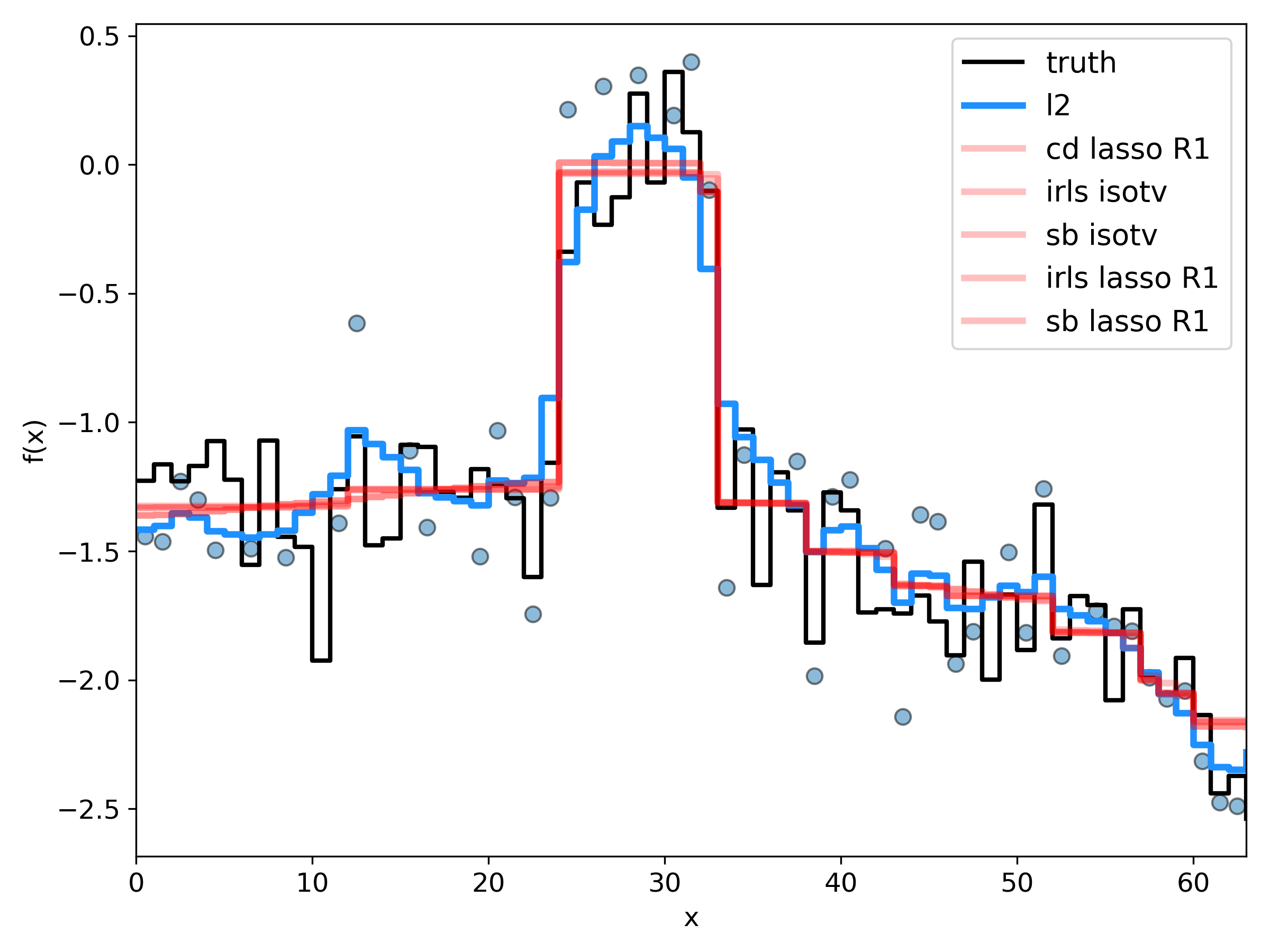}
	\includegraphics[width=0.5\linewidth]{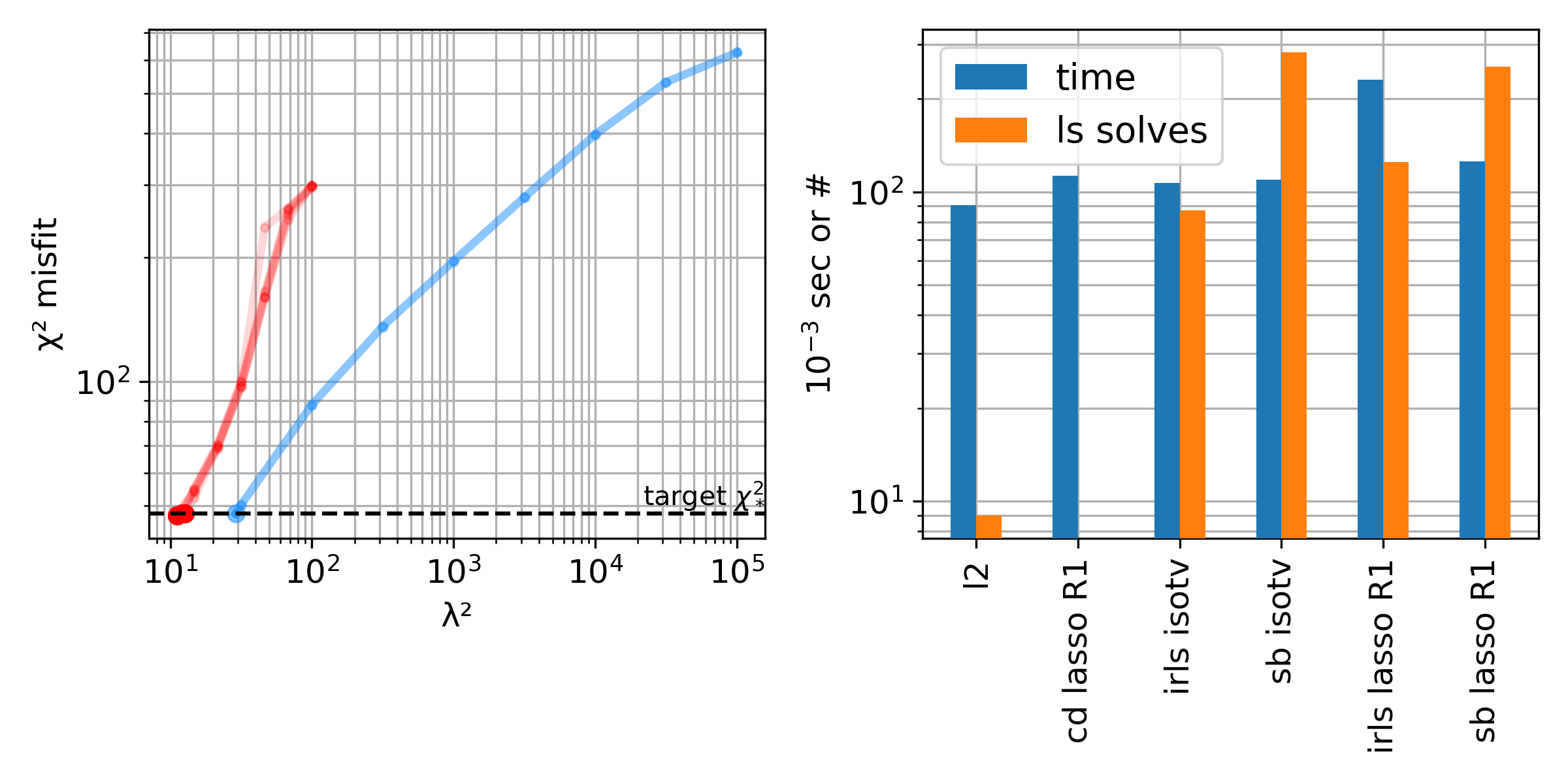}
	\caption{The top row shows the true model (black line), noisy data samples (outlined blue circles), the $\ell_2$ reconstruction (blue line), and various $\ell_1$ reconstructions (red lines), converging on largely the same model. The true model is based on natural property variations in the earth, which we will revisit in the nonlinear problem in Section~\ref{sec:aem}. The left bottom panel shows each regularisation path for the $\ell_2$ and $\ell_1$ problems. The right bottom panel shows the time taken in milliseconds as well as the final number of least squares solves for each method. As expected, the $\ell_1$ regularisation paths are steeper than $\ell_2$, and it is remarkable that the $\ell_1$ solutions and regularisation paths are quite similar, despite algorithmic differences between them (see text for details). Also as expected, the near-identical $\ell_1$ fused lasso solutions are blockier, despite reaching the same misfit target as the $\ell_2$ solution.} 
	\label{fig:compare1d}
\end{figure}
This is a linear problem of the form \eqref{eq:fused_lasso_shifted} for $\ell_1$ regularisation and the linear equivalent of \eqref{eqn:phi} for conventional $\ell_2$ regularisation. The forward matrix is a simple restriction operator, ie., a row-rank deficient subsampling matrix $\mathbf S$ which acts on a vector $\mathbf m$. For each observed pixel corresponding to a row, there is a one in the corresponding column for the retained pixel, and zeros elsewhere. For example, to select the second element in column vector $\mathbf m$, we act on it from the left with the unit basis given by
$
\mathbf e_2 = 
\begin{bmatrix}
0 &1&\cdots&0
\end{bmatrix}
$.
Explicitly, we write the forward problem as follows:
\begin{align}
\mathbf y &= \mathbf{Sm}, \quad \mathbf{S} \in\mathbb{R}^{n_d\times n_p},\\
\begin{bmatrix}
\vdots\\
y_i\\
\vdots\\
\end{bmatrix} &= 
\begin{bmatrix}
\vdots\\
\mathbf e_{j(i)}\\
\vdots\\
\end{bmatrix} 
\begin{bmatrix}
m_1\\
\vdots\\
m_{n_p} 
\end{bmatrix}, \quad j(i)\in \{1,\cdots,n_p\}; \; i \leq n_d < n_p,
\end{align}
where $\mathbf e_{j(i)}$ picks out the $j^\text{th}$ element of $\mathbf m$ for the $i^\text{th}$ observation. The data in $\mathbf y$ are contaminated with homoskedastic Gaussian noise of standard deviation 0.25, with $n_d = 48$ out of $n_p = 64$ observation points randomly retained. The inverse problem with known noise level is to reconstruct the true vector $\mathbf m$ (Figure~\ref{fig:compare1d}), with a target $\chi^2_*=48$. The true model we should note, does not play exclusively to the strengths of either $\ell_1$ or $\ell_2$ regularisation. For the $\ell_1$ problem with a non zero $\beta^2=0.0005$ and reference prior model set to $\mathbf{-2}$, we remind the reader that the solution methodology is to first solve \eqref{eq:z_lasso_problem} followed by soft thresholding with \eqref{eq:soft_threshold_scalar}. In 1D the analysis mode isotropic TV formulations \eqref{eq:cgls_stacked} and \eqref{eq:sb_tv} are equivalent to the invertible form lasso problem \eqref{eq:z_lasso_problem}. The isotropic TV problem can be solved using split Bregman and IRLS. On the other hand, the lasso problem can be natively solved with coordinate descent, as well as with split Bregman and IRLS. This presents an ideal opportunity to test the different methods and our implementation of IRLS and split Bregman, noting that we use the Lasso.jl coordinate descent algorithm method based on \cite{friedman2010regularization}. This is because all algorithms should produce virtually the same solution along very similar regularisation paths, given their mathematical equivalence for a linear, convex problem. With a relative tolerance of $10^{-2}$ for both split Bregman and IRLS, we see that this is indeed the case in Figure~\ref{fig:compare1d}. The coordinate descent algorithm used an absolute value tolerance of $10^{-6}$ between lasso weights in successive iterations as a stopping criterion. Tightening all tolerances by an order of magnitude produced similar results to Figure~\ref{fig:compare1d}, but with even less spread. There were no discernible differences between $\ell_1$ solutions and their regularisation paths, at the cost of many hundreds of least squares solves. We note at this point, that the split Bregman method requires 2-3 times the number of least squares solves as does the IRLS method for the same problem. This is something to keep in mind as we approach the 2D problem next, with an added layer of complexity and of course, an increase in dimensionality. Both the $\ell_2$ and $\ell_1$ solutions achieve the target misfit, but as expected, the $\ell_1$ solutions are blockier. 
\subsection{A linear 2D problem: image denoising and deblurring}
\begin{figure}
	\centering
	\includegraphics[width=0.7\linewidth]{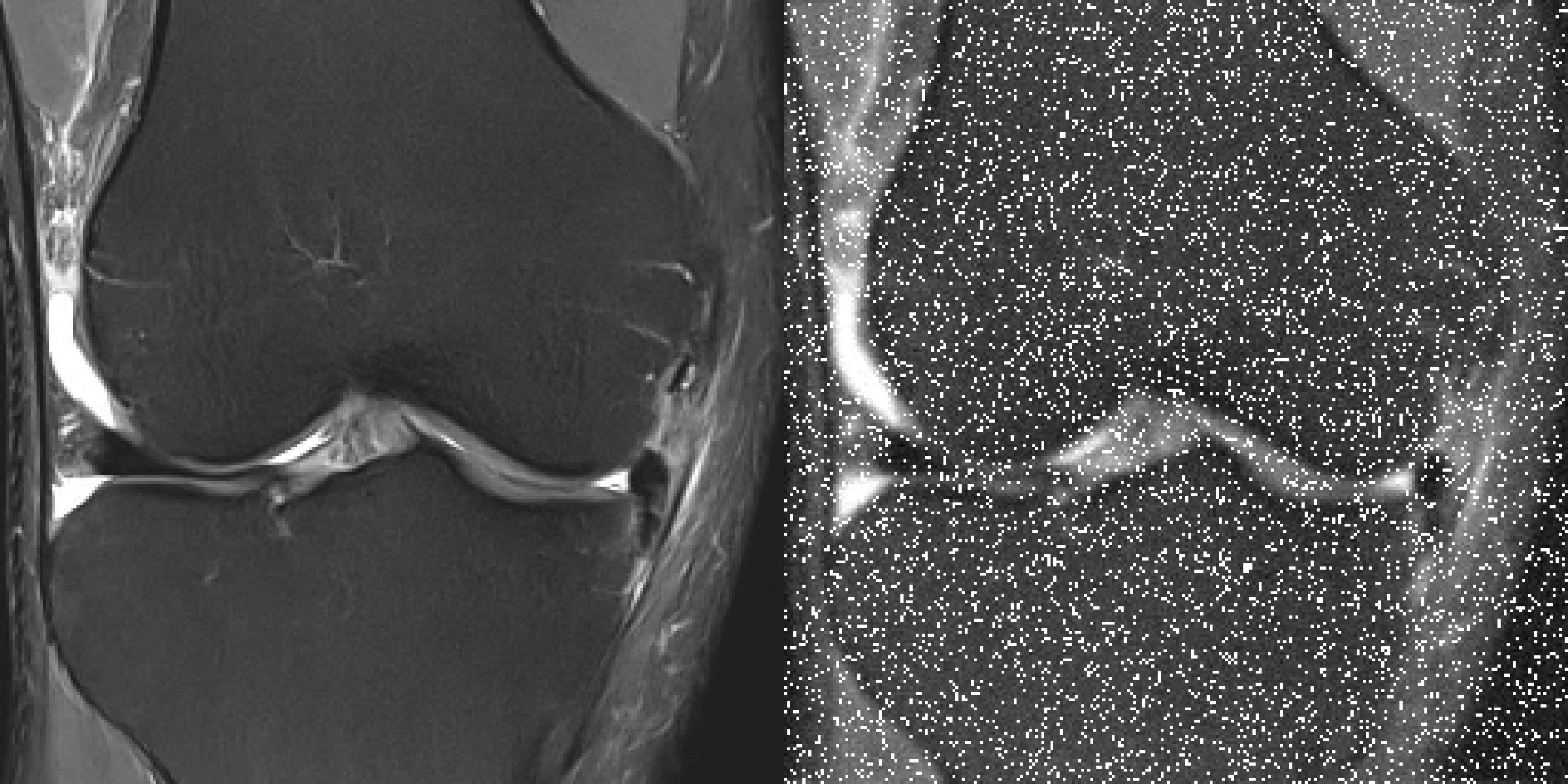}
	\caption{The left panel shows a $256 \times 256$ true MRI slice through the right knee joint (front facing, coronal plane). The right panel shows the same image, but with a 10-pixel Butterworth low pass applied along both the rows and columns, 3\% noise added and 90\% of pixels retained. The inverse problem is to reconstruct the original image from the noisy samples.} 
	\label{fig:2Dproblem}
\end{figure}
\noindent 
We now consider another linear problem, but with added complexity to what we have just analysed in 1D. Given a high fidelity image, e.g., a magnetic resonance imaging (MRI) medical scan, the image is low pass filtered along each coordinate axis, then subsampled, and Gaussian noise is finally added to it, forming the final data $\mathbf y$. The inverse problem is to reconstruct the original image by undoing the effects of the blur and subsampling, when the data noise level is known. Reusing the subsampling operator $\mathbf{S}$ from the last section, writing the Fourier transform as $\mathbf{F}$, the diagonal Fourier-domain Butterworth filter as $\mathbf{L}$, and denoting conjugate transposition with $^\dagger$, the forward problem can be written using \eqref{eq:2D_kron_D} as:
\begin{align}
\mathbf y &= \mathbf{S(F}_2^\dagger \mathbf{L}_2 \mathbf{F}_2) \otimes 
	(\mathbf{F}^\dagger_1 \mathbf{L}_1 \mathbf{F}_1)\mathbf m,\\
		&=\mathbf{S} \mathbf{B}_2 \otimes \mathbf{B}_1 \mathbf{m},
\end{align}
where $\mathbf{B}_i$ is the space domain Butterworth blurring operator in dimension $i$.
\begin{figure}
	\centering
	\includegraphics[width=0.7\linewidth]{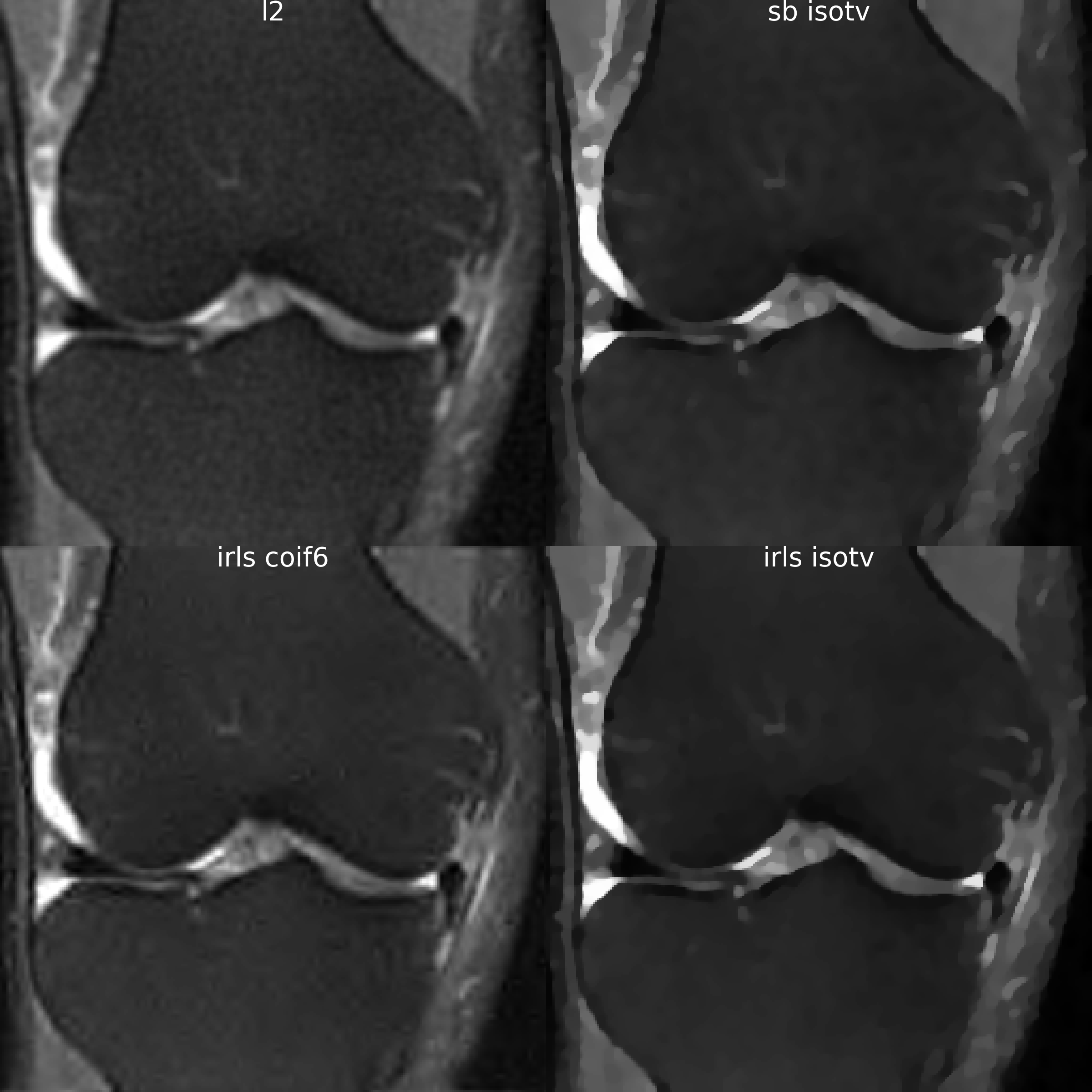}
	\caption{Deblurred imaged reconstructions. The top left panel shows the $\ell_2$ least squares reconstruction. The bottom left panel shows the $\ell_1$ reconstruction using the Coiflet-6 basis family. Both figures on the right are $\ell_1$ isotropic TV reconstructions using either split Bregman (SB) or IRLS. As expected, the isotropic TV reconstruction shows better segmenting of the various zones within the joint. Some quantitative metrics for comparison are provided in Table~\ref{tab:imagemetrics}.} 
	\label{fig:2Drecon}
\end{figure}
\begin{figure}
	\centering
	\includegraphics[width=0.5\linewidth]{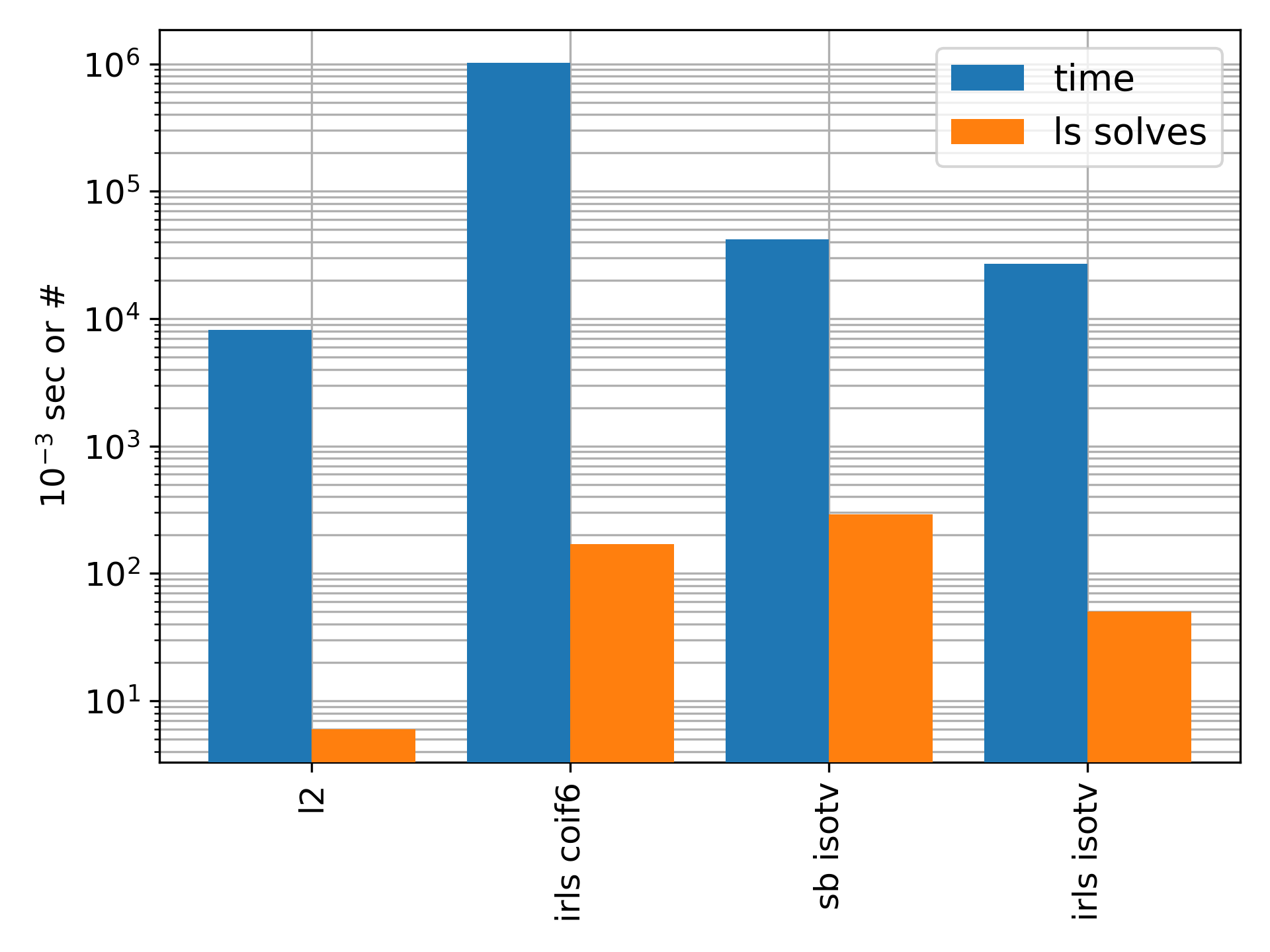}
	\caption{Following a pattern similar to what was observed for the 1D linear problem, we see that completion time increases with the number of least squares solves for the split Bregman and IRLS isotropic TV solutions. For isotropic TV, the split Bregman method requires nearly 60\% more time than IRLS, owing to nearly 6 times as many least squares solves. The greatest amount of time is taken by the Coiflet solution, despite a smaller number of least squares solves than split Bregman. This is due to the fact that the solution machinery was not memory preallocated and optimised, unlike the optimised gradient operators used for isotropic TV.} 
	\label{fig:2Dtimings}
\end{figure}
\begin{figure}
	\centering
	\includegraphics[width=0.7\linewidth]{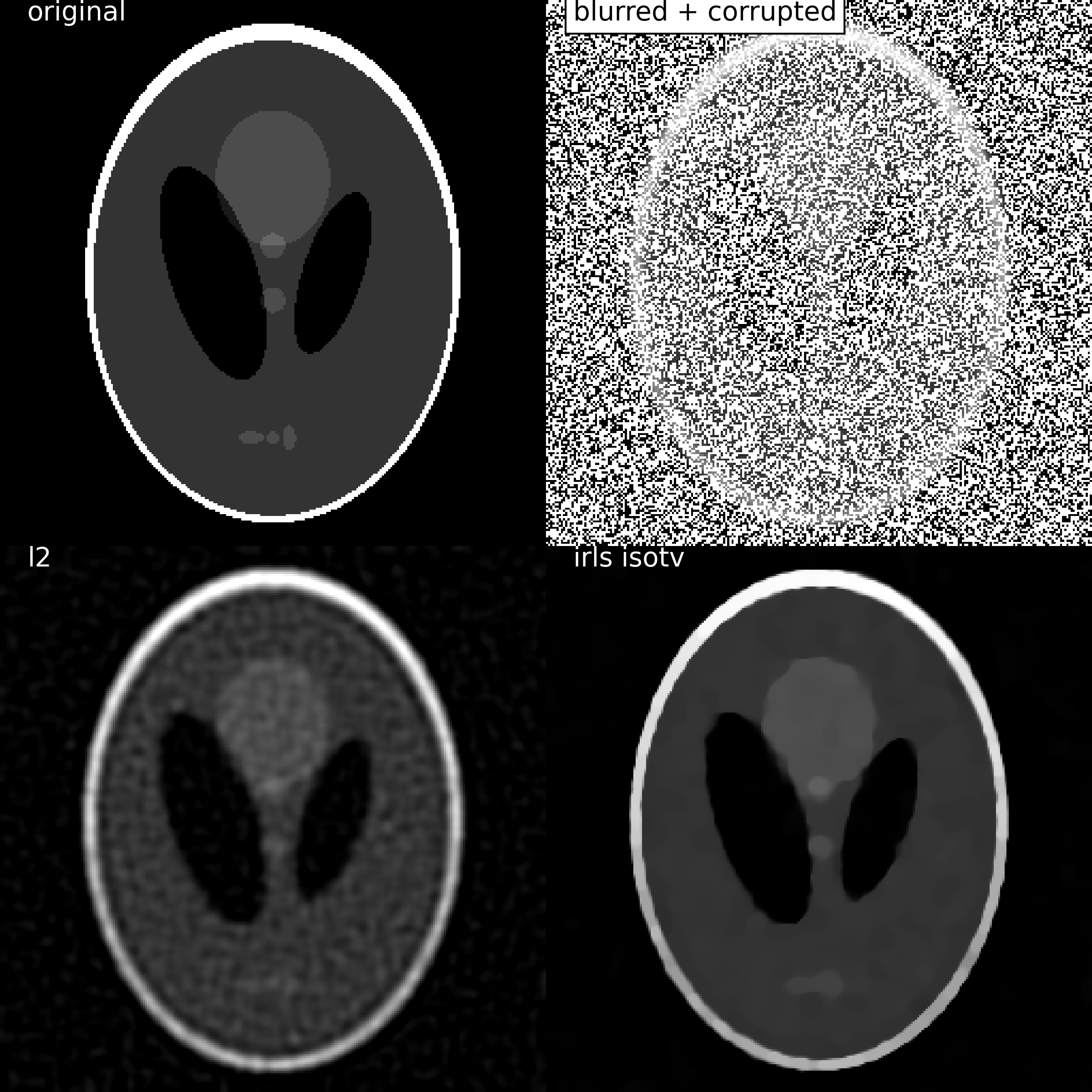}
	\caption{The $256\times256$ Shepp-Logan phantom test. The true image is in the top left, and its blurred, subsampled and noisy version with only 40\% of data retained with 3\% noise added is in the top right. The blur applied is a 20 pixel Butterworth along both the rows and columns. The bottom left image is the $\ell_2$ least squares reconstruction. The bottom right panel shows the $\ell_1$ isotropic TV reconstruction using IRLS. Clearly, the isotropic TV reconstruction shows better reconstruction of the original. Some quantitative metrics for comparison are provided in Table~\ref{tab:imagemetrics2}.} 
	\label{fig:phantom}
\end{figure}
\begin{table}
{
\caption{Quantitative $\chi^2$ misfit and SSIM metrics for the knee joint MRI slice reconstruction. Higher SSIM values imply better visual reconstruction. Both $\ell_2$ and $\ell_1$ fit the observations almost exactly to within noise, but the $\ell_1$ cases have a better SSIM.}
\begin{center}
	\begin{tabular}{l|r|r}
     \hline 
     Regularisation type & $\chi^2$ achieved (target=58, 982) & SSIM\\
     \hline
    $\ell_2$ & 58, 982 & 0.80\\
     \hline
    IRLS Coiflet-6 & 58, 977 & 0.81\\
     \hline
     SB Iso-TV & 58, 944 & 0.83\\
    \hline
    IRLS Iso-TV & 58, 978 &0.83\\
		\end{tabular}
	\end{center}
\label{tab:imagemetrics}}
\end{table}
\begin{table}
{
\caption{Quantitative $\chi^2$ and SSIM metrics for the Shepp-Logan phantom reconstruction. The $\ell_2$ reconstruction does not completely fit the data, and the SSIM is significantly lower than the $\ell_1$ reconstruction.}
\begin{center}
	\begin{tabular}{l|r|r}
     \hline 
     Regularisation type & $\chi^2$ achieved (target=26, 214) & SSIM\\
     \hline
    $\ell_2$ & 29,077 & 0.60\\
    \hline
    IRLS Iso-TV & 26,212 &0.90\\
		\end{tabular}
	\end{center}
\label{tab:imagemetrics2}
}
\end{table}
In this example, given the large areas of little variation in Figure~\ref{fig:2Dproblem} coupled with sharp edges, we would expect the isotropic TV solution to perform somewhat better than the $\ell_2$ solution. We qualify our stated expectations with the word somewhat because there are textural components in the original knee image which TV will not capture well. A comparison between three regularisation schemes -- $\ell_2$ using \eqref{eqn:phi}, $\ell_1$ and the wavelet transform with the orthogonal Coiflet basis \citep[see, e.g.,][]{graps1995introduction} using \eqref{eq:z_lasso_problem}, and analysis mode $\ell_1$ isotropic TV using  IRLS \eqref{eq:cgls_stacked} and split Bregman \eqref{eq:sb} are shown in Figure~\ref{fig:2Drecon}. All experiments fit the data to within noise, and our expectations are met, as confirmed in Table~\ref{tab:imagemetrics} using a metric known as the structural similarity index metric  \citep[SSIM, see][]{wang2004image}. The index ranges from 0 to 1 and provides a good measure of image restoration as perceived by human vision.  

For this image with 65,536 pixels, the time taken is longer than for the 64 pixel 1D problem previously considered. The number of least squares solves for the IRLS and split Bregman methods increases considerably. Further, we were no longer able to use the coordinate descent method for the lasso Coiflet problem due to matrix materialisation costs in the Lasso.jl coordinate descent algorithm, and had to switch to our IRLS implementation that only uses forward operators and their adjoints with no explicit materialisation. The timing and number of solves are graphically shown in Figure~\ref{fig:2Dtimings}. Just as in the 1D problem, the IRLS and split Bregman problems arrive at virtually identical iso TV solutions, but IRLS requires 26 seconds and 50 total least squares solves across the $\lambda^2$ sweep to convergence within noise (including root finding), whereas the split Bregman method required 42 seconds and 290 least squares solves. We surmise that this is due to the selection of $\lambda^2=\gamma$, which controls the rate of convergence to the true solution. While alternatives are provided in \cite{boyd2011distributed}, we do not investigate them here due to the satisfactory performance of the IRLS method. This also brings us to an interesting juncture from which to view the ill-conditioned nonlinear problem, for which there is no guarantee of descent to a unique minimum, if the descent trajectory differs in the initial stages of the optimisation problem. Before we move on to the nonlinear problem however, it is important to understand the role of the prior, and how dramatically this can change the reconstruction, and hence the interpretation of images.

If the image is indeed sparse in a particular reconstruction basis, for instance if it were to have clear edges, little to no texture, and large flat zones of little intensity variation, which would make it an ideal target for isotropic TV -- we should theoretically be able to reconstruct it well from very few samples. To demonstrate this phenomenon, we use the famous Shepp-Logan phantom image \citep{shepp1974fourier}, which has been used in numerous studies to demonstrate sparsity promoting imaging techniques \citep[e.g.,][]{candes2006robust} in the same deblurring context as considered previously. The results for the $256\times256$ image, but this time with only 40\% of pixels retained, are shown in Figure~\ref{fig:phantom}. The $\ell_2$ inversion does not quite fit the data to within noise ($\chi^2 > \chi^2_*$), while the $\ell_1$ isotropic TV inversion fits the data to within noise and is visually a better reconstruction, as borne out by a 50\% higher SSIM value (Table~\ref{tab:imagemetrics2}). Contrasting these two examples, we see that prior assumptions make a significant difference in the recovery of an underlying model. If the data are uninformative, or if the true model aligns with the prior (sparsity promoting in this case), then the solution is truly sparse (aligned with the prior). In the nonlinear problem considered next, we are in a situation that lies somewhere in-between these two extremes.
\subsection{A synthetic nonlinear problem: transient airborne electromagnetic sounding}
\label{sec:aem}
\begin{figure}
	\centering
	\includegraphics[width=1\linewidth]{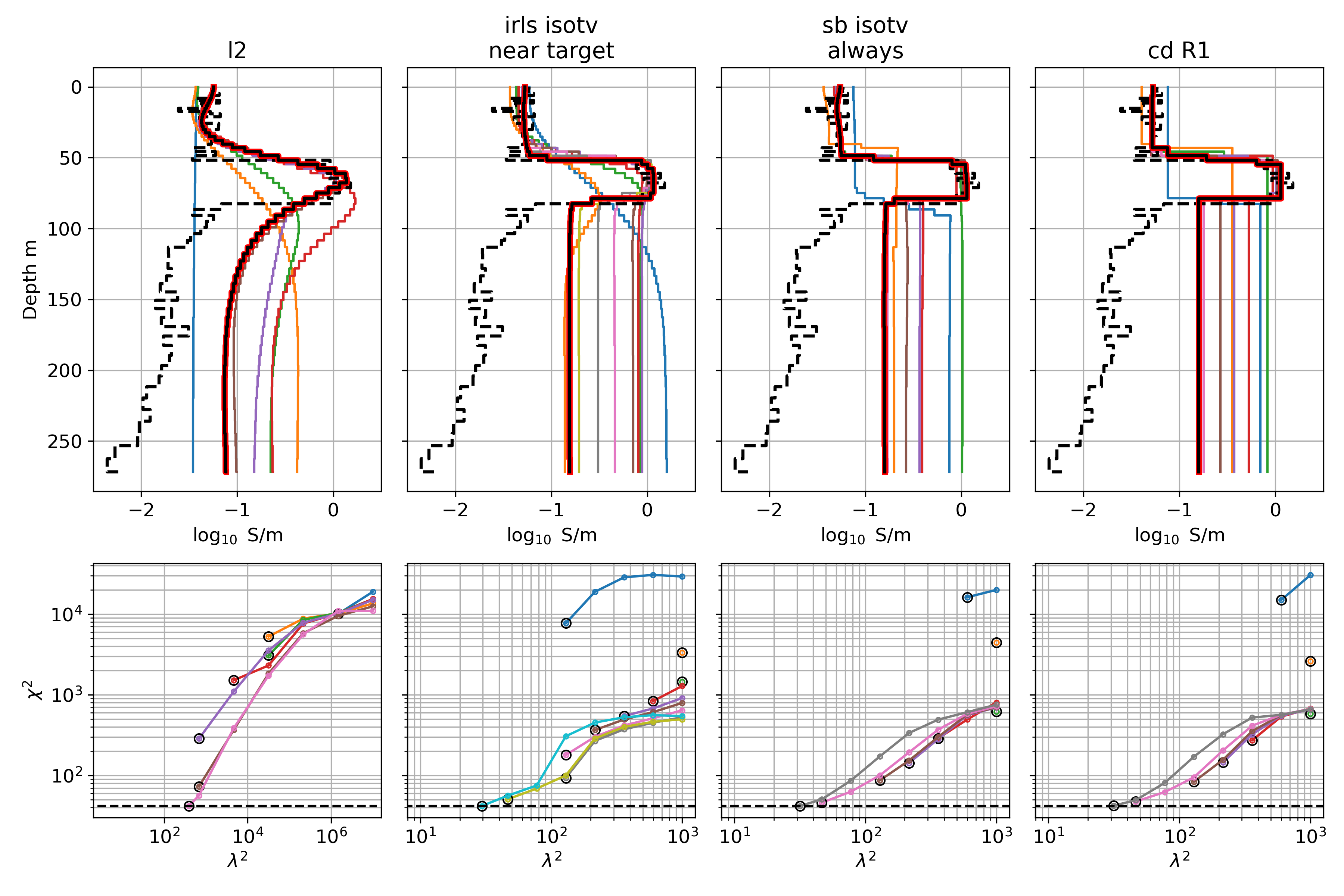}
	\caption{The nonlinear AEM problem, with each column representing the solution method. The top row shows the best solutions at each linearisation, with the true model provided in dashed black, and the thick red and black line showing the final solution. The bottom row shows the regularisation path during each nonlinear iteration (i.e., a fixed linearisation), the selected $\lambda^2$ is circled in black, and the dashed line indicates the target $\chi^2$. Again, the $\ell_2$ solution is smoother than all the $\ell_1$ solutions. The $\ell_1$ solutions are not interpretably different from a geological point of view. All methods are tending to the prior of $\log_{10} 0.01$ S/m in a norm-specific manner and have limited or no sensitivity to the resistive-trending structure below 100 m. The 3-layered background-saline aquifer-background structure is again, most starkly highlighted by the $\ell_1$ solutions.} 
	\label{fig:AEMproblem}
\end{figure}
\begin{figure}
	\centering
	\includegraphics[width=0.5\linewidth]{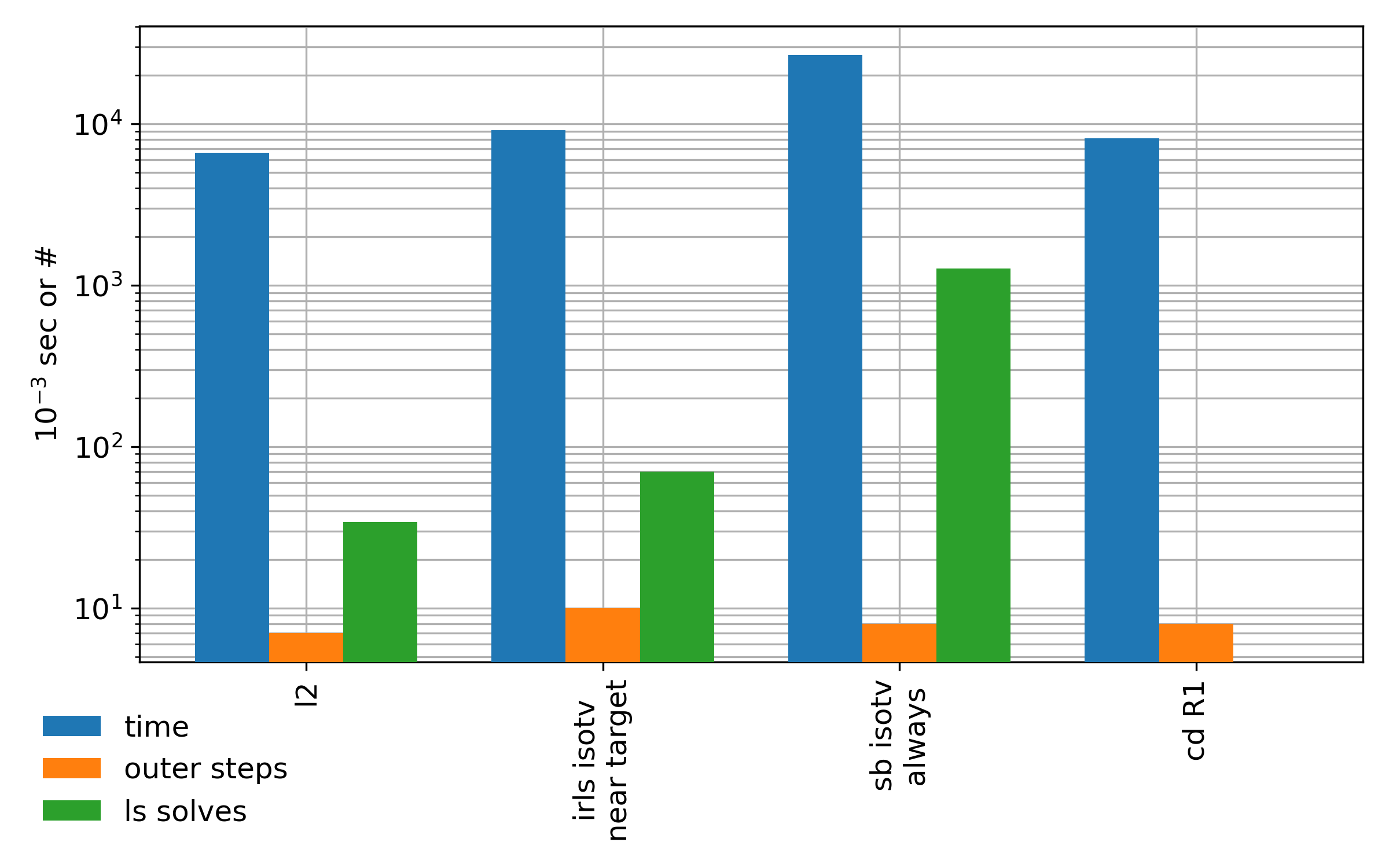}
	\caption{Following a pattern similar to what was observed for both the linear problems, we see that completion time for the nonlinear AEM problem increases with the number of least squares solves, for the split Bregman and IRLS isotropic TV solutions. The split Bregman method was nearly 3 times slower than IRLS, owing to an order of magnitude more least squares solves. As in the 1D linear problem, coordinate descent required time comparable to the $\ell_2$ solution, and importantly, no least squares solves.} 
	\label{fig:compareAEM}
\end{figure}
The nonlinear AEM inverse problem, is of the form 
\begin{equation}
	\mathbf{d} = \mathbf{f(m)}, \label{eq:nonlinfwd}
\end{equation}
where $\mathbf m$ is the subsurface $\log_{10}$ conductivity with depth. $\mathbf {f(m)}$ in the problem considered, is the nonlinear map from subsurface conductivity to the time domain transient magnetic field calculated at receiver locations. Specifically, $\mathbf {f(m)}$ provides time derivatives of the transients proportional to voltage, for a helicopter AEM system, exactly as described in \cite{Ray2023}. The AEM method is preferentially sensitive to conductive subsurface geology, and has extensively been used for a host of geophysical reconnaissance and mapping purposes \citep[see Chapter 1,][for a concise overview of the method]{Brodie2010}. A minimum gradient support \citep{portniaguine1999focusing} inversion methodology that produces models with sharp discontinuities has been used for the AEM problem \citep{vignoli2015sharp}, however the regularisation scheme is an $\ell_0$ approximator, and differs from the convex $\ell_1$ family discussed in this work.

In Section~\ref{sec:nonlinear}, we showed that regularised solution of \eqref{eq:nonlinfwd} requires linearisation of the form~\eqref{eq:linearised_objective_linear_form} and the solution of the linear problem in a sequence of outer steps, until the target misfit is exceeded. As recommended by \cite{Brodie2010} and implemented in \cite{Ray2026}, for the AEM problem, we also cut short the sweep across $\lambda^2$ during any one linearisation, if we descend to 70\% or lower of the lowest misfit in the previous step. Other practitioners in the electromagnetics literature have recommended similar descent strategies for the nonlinear induction problem \citep[e.g.,][]{Farquharson1993, Key2016}, which ensures that we do not head in the wrong descent direction too early in the nonlinear problem. If in any outer step, no value of $\lambda^2$ yields a lower misfit than the previous step, a convex combination of the proposed model and the previous model equivalent to halving the step size is carried out as described by both \cite{DeGroot-Hedlin2004} and \cite{Key2016}. Following the descent phase of Occam as described in the original \cite{Constable1987} paper, bracketing is then carried out to find the optimal $\lambda^2_*$ that yields the target $\chi^2_*$ with our secant modifications as described in Section~\ref{sec:phase2}. 

The results of a synthetic inversion with data contaminated with 3\% Gaussian noise are shown in Figure~\ref{fig:AEMproblem}. We have taken $f(x)$ from Figure~\ref{fig:compare1d}, assigned it units of $\log_{10}$ S/m and arranged it as a function of depth, to represent our true model of conductivity within the earth. Details of the synthetic model are exactly the same as for the helicopter AEM synthetic in \cite{Ray2023} and are not repeated here. Just as in the linear 1D problem in Section~\ref{sec:1Dproblem}, $\beta^2=0.0005$ and the reference prior model is set to $\mathbf{-2}$. Comparing the $\ell_2$ solution and regularisation path with $\ell_1$ isotropic TV solutions and regularisation paths found using IRLS, split Bregman, and coordinate descent, we note the following. As expected, the $\ell_2$ solution is smoothly varying and does not capture the aquifer geometry as well as the $\ell_1$ solutions do. This is yet another example of how the prior assumptions ($\ell_2$ vs $\ell_1$ regularisation) make for a considerable difference in the character of the final solution. The regularisation paths show that we could probably implement a ``fast-Occam'' acceleration where we start each linearisation's $\lambda^2$ sweep from the best fitting $\lambda^2$ in the previous linearisation's sweep \citep{Key2016}. All the final $\ell_1$ solutions are relatively similar, though the descent paths are quite different for IRLS. As described in Section~\ref{sec:sb_relation} contrasting split Bregman and IRLS, the left and right hand sides of the least squares parts (and hence the linearised descent trajectory) of each method are handled differently. For the nonlinear AEM problem, we found that running IRLS reweighting early, leads to too much spread in the early descent stage models. This can be explained as follows: As the algorithm proceeds, during each linear sweep over $\lambda^2$, the model (and hence the IRLS weights, which are computed from the model's spatial gradient at the previously visited value of $\lambda^2$) converges towards the total variation minimum at that stage of the descent. It is wasteful and potentially harmful to iterate excessively towards an initially blocky but incorrect model that the Jacobian at that point cannot provide a path away from. This risks pushing descent towards the wrong minimum due to column space modifications of the least squares inverse in~\eqref{eq:cgls_stacked}. At later stages, when we are in the vicinity of the total variation minimum that we are aiming for (i.e., ``near target'' in Figure~\ref{fig:AEMproblem}), multiple weight refinement steps at each value of $\lambda^2$ are run. ``Near target'' was defined as 4 times the target $\chi^2$ of 42 for this example. Conversely the split Bregman method required Bregman updates for the $\ell_1$ constraints to be met right from the start of the descent. This contrasting behaviour to IRLS can be understood as follows: The Bregman variable accumulation~\eqref{eq:bregmanacc} accommodates the shift in the Lagrange multiplier from~\eqref{eq:stationarity_unaugmented} to~\eqref{eq:stationarity_augmented}. Since the Bregman and auxiliary variables $\mathbf{b, d}$ are reset \textit{every} $\lambda^2$, there is no memory from a previous $\lambda^2$ that can assist in convergence to the total variation minimum for any stage of linearisation. Enough Bregman iterations~\eqref{eq:sb_stacked}--\eqref{eq:sb_update} need to be run such that these quantities (and hence the iterated model) stabilises. The coordinate descent method used the same absolute tolerance of $10^{-6}$ of change in $\mathbf m$ as before. 

Figure~\ref{fig:compareAEM} contrasts the time taken and least squares solves required by each method. The number of IRLS weight refinement or split Bregman updating least squares solves for each method was limited to 50 (per $\lambda^2$), yet across 10 and 8 outer iterations respectively, IRLS required a total of 70 solves and split Bregman 1268 total solves. This is reflected in the amount of time taken by each, with IRLS taking 9.1 seconds, to split Bregman's 26.1 seconds. This is again, we surmise, due to the selection of $\gamma = \lambda^2$, as pointed out in the 2D deblurring problem. Perhaps surprisingly, with no least squares solves, the coordinate descent method of \cite{friedman2010regularization} is quite compute-wise efficient, as it was for the linear 1D problem. As we do not control the internals of the Lasso.jl coordinate descent algorithm, we were unable to intercept the coordinate descent path near the root finding stage and hence did not use the ``pathwise'' coordinate descent method, and carried out a separate solve per $\lambda^2$. Had we  carried out coordinate descent over  the entire $\lambda^2$ sequence as described in \cite{friedman2010regularization} similar to our ``near target'' IRLS scheme, it would have been the fastest method for this problem.
\subsection{Field data: transient airborne electromagnetic imaging}
\begin{table}
{
\caption{AEM field data inversion statistics}
\begin{center}
	\begin{tabular}{l|r|r|r}
     %\hline 
     Inversion type & soundings with RMS$<1.05$ & max RMS & wall clock time\\
     \hline
    $\ell_2$ & 99.9\% & 1.46 & 92 s\\
     \hline
    $\ell_1$ fused lasso & 98.8\% & 1.51 & 99 s \\
     		\end{tabular}
	\end{center}
\label{tab:fieldmetrics}}
\end{table}
The final example draws on what we have learned from the linear and nonlinear examples studied so far. Instead of a single AEM sounding, 720 soundings from a field helicopter AEM survey line are inverted sounding-by-sounding to generate a conductivity-depth section of the earth. The data are from Geoscience Australia's test range in Menindee, New South Wales, Australia. Various AEM imaging systems have been extensively tested on a particular flightline across Menindee lake \citep[see][for details]{Ray2026}. The particular dataset used here corresponds to the 2015 SkyTEM survey, the details of which can be found in the aforementioned reference. The same AEM system was previously studied in the nonlinear synthetic study of Section~\ref{sec:aem}.  Accurate geophysical noise estimates are crucial for Occam's inversion. Details of the noise estimation at the test range for helicopter systems can be found in \cite{scarr2026estimating} and are not discussed here.
\begin{figure}
	\centering
	\includegraphics[width=1\linewidth]{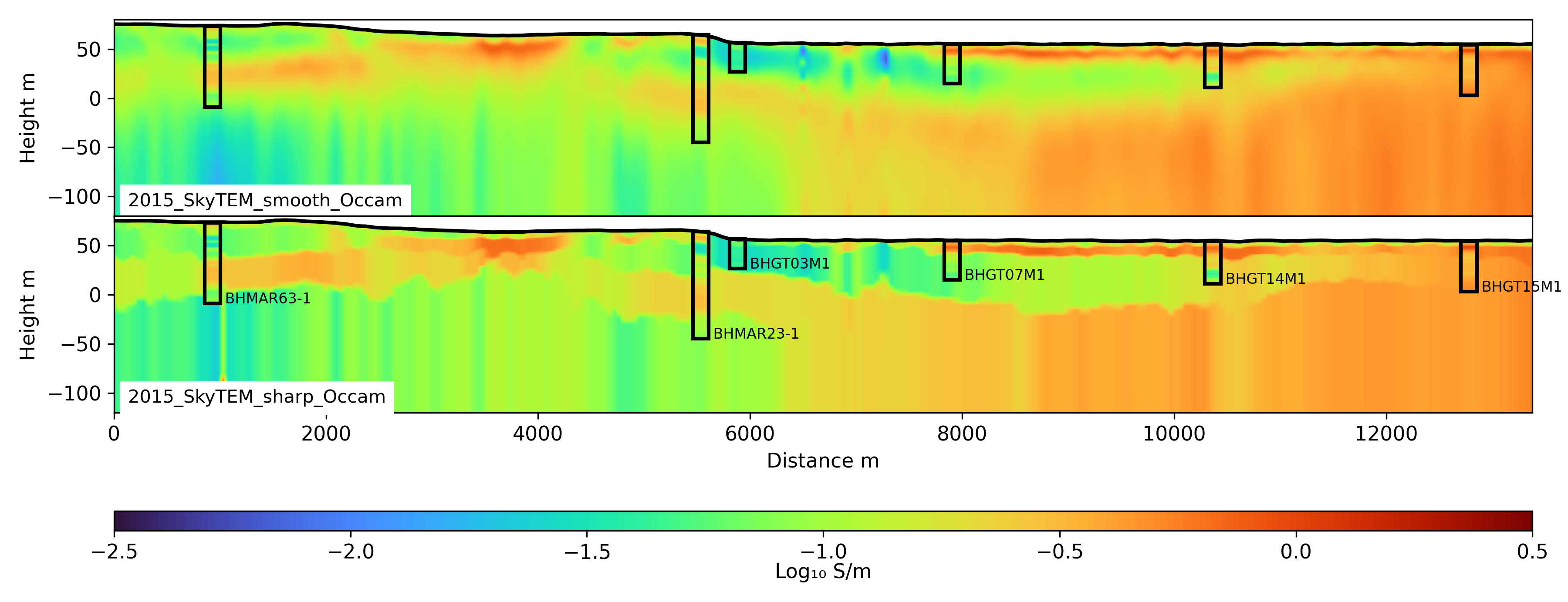}
	\caption{Occam's $\ell_2$ traditional inversion (top) vs Occam's $\ell_1$-TV inversion with coordinate descent (bottom) using SkyTEM 312 data (2015) from the Menindee test range. From a geological perspective, the bottom image is easier to interpret -- as in the 2D deblurring case, it is better segmented. Some statistics are provided in Table~\ref{tab:fieldmetrics}.} 
	\label{fig:compareAEMfield}
\end{figure}

The inversion results comparing $\ell_2$ and $\ell_1$ regularisation are shown in Figure~\ref{fig:compareAEMfield}. It is remarkable that with both inversion types, sounding-by-sounding inversion preserves lateral continuity. In terms of differences between the smooth and sharp inversions, the observations from the preceding example still hold true. While both inversions show good alignment with known geology and the borehole induction log data (the closest analogue to ground truth in geophysics), sharp conductivity contrasts are recovered with the $\ell_1$ inversion. As expected, the $\ell_1$ inversion displays minimal oscillatory ``wiggliness" in the vertical direction. Consequently, geological interpretation can be more easily carried out with the $\ell_1$ sharp inversion compared to the standard $\ell_2$ smooth inversion. 

As expected from the synthetic AEM sounding study in the preceding section (Section~\ref{sec:aem}), the $\ell_1$ inversions required similar times to the $\ell_2$ inversions, and coordinate descent was used as the default solver for the 1D fused lasso problem with $\beta^2=0.0005$, the same as in both the linear and nonlinear 1D synthetic examples. Both inversions were inverted on a high performance compute cluster, across 720 CPUs in parallel. With a maximum of 40 outer iterations, the entire line was inverted in 92 s for the $\ell_2$ inversions and 99 s for $\ell_1$. As is often the case for electromagnetic geophysics, instead of reporting the $\chi^2$ misfit achieved, the root mean square (RMS) error obtained by calculating $\sqrt{\frac{\chi^2}{n_\text{data}}}$ is mentioned. As explained earlier in Section~\ref{sec:phase2}, inverted conductivity models that fit the data well will achieve an RMS misfit $\approx 1$. Keeping this in mind, for the $\ell_2$ inversions, 719 out of 720 inversions fit the data to within an RMS misfit of 1.05, with one sounding fitting to an RMS of 1.46. For the $\ell_1$ inversions, the data fit statistics were quite similar, with 711 out of 720 soundings fitting to within an RMS misfit of 1.05, and the maximum RMS misfit was 1.51. These values are summarised in Table~\ref{tab:fieldmetrics}. From a practical standpoint, all the data were fit to within the data noise by both inversion types. Most importantly, this was achieved within a comparable timeframe, using coordinate descent (followed by soft-thresholding shrinkage to the reference model) for the $\ell_1$ regularisation.
\section{Discussion and conclusions}
We have provided a unified framework for invertible, synthesis and analysis forms of $\ell_1$ regularised inverse problems within the Occam framework, for both linear and nonlinear problems. To encourage experimentation and uptake of the various algorithms required to enforce $\ell_1$ regularisation, we have elucidated the mathematics behind lasso fusion, total variation in multiple dimensions, and implementation using the coordinate descent, IRLS and split Bregman methods. We have also derived split Bregman in the light of ADMM and Lagrange multipliers that are familiar to geophysicists. We have also shown the equivalence of these algorithms through a sequence of linear and nonlinear problems, culminating in a field data example involving airborne transient electromagnetics over the Menindee test range in Australia. We are not aware of previous implementations of fused lasso coordinate descent for the AEM problem, and have found it to be a surprisingly effective algorithm for small to medium scale 1D problems. For larger problems in higher dimensions, we show that the IRLS algorithm is quite effective, and that the split Bregman problem with the default choices of regularisation parameter is slow to converge. From our studies of the nonlinear AEM problem, we recommend running multiple IRLS reweighting steps when we are near the misfit target. This is a promising way to economise the number of time consuming least squares solves. In this spirit, running a number of secant steps with multiple reweighting passes in phase II of the Occam inversion seems to provide a stable means of converging to a desirable misfit minimum, while ensuring sharpness of model discontinuities. 

Through our deblurring examples involving a knee MRI image and the Shepp-Logan phantom, we show clearly how the choice of prior assumptions (smooth vs sharp) and the nature of the true model itself inform the outcome of the reconstructed image after inversion. Interestingly, we did not encounter in our data decimation experiments down to 10\% for the Shepp-Logan phantom, the sudden threshold between ``poor'' and ``excellent'' $\ell_1$ recovery \citep[see][]{candes2008enhancing}. This absence of threshold is also reported by \cite{turuncctur2023overcomplete} when dealing with noisy data. We surmise this is because we were working with noisy data as well as operating within the Occam framework of ``simple but no simpler". One powerful solver we did not test was fast iterative shrinkage-thresholding \citep[FISTA,][]{beck2009fast}, which is naturally suited to the synthesis form of the $\ell_1$ regularised problem. Future directions could involve investigating spatially variable norms \citep{fournier2019} or the more general elastic-net which combines the best features of both $\ell_1$ and $\ell_2$ regularisation \citep{zouelastic}. With IRLS and modifications to majorisers of the form \eqref{eqn:surrogate} and~\eqref{eqn:eta2} the entire Occam framework for $\ell_p$ with $0<p<1$ can also be handled for more sparsity than provided by $p=1$, but at the cost of the convexity guarantee that $\ell_1$ provides \citep{daubechies2010iteratively, mairal2014sparse, kaushik2024precise}. We stress that no inversion scheme is universally better. There are always tradeoffs regarding algorithm complexity, information content in the observations, scalability of the algorithm, and compatibility of the assumptions made with the actual geophysical processes that drive the true model under investigation. 

\section*{Acknowledgments}
I thank Michael Koch and Ross Brodie for useful conversations on inverse problems and improving their regularisation. Malcolm Sambridge, Sam Kaplan and John Washbourne introduced me to basis pursuit and $l_1$ methods. I am indebted to Steven Constable, Catherine Constable, and Kerry Key for showing me the inner workings of Occam's inversion. Roger Miller encouraged me to pursue the majorisation-minimisation route. Mona Mahani and Tim Scarr provided useful initial reviews and feedback. This study was funded by the Australian Government's Resourcing Australia's Prosperity initiative, with a view to increasing precompetitive geoscience knowledge. Test range data were inverted on the \textit{Gadi} high performance computing cluster at the National Computational Infrastructure, at the Australian National University, Canberra.  

\section*{Data availability}
All computation was carried out using the open source Julia language \citep{Bezanson2017} v1.12.4 and by modifying the HiQGA v0.5.0 package \citep{Ray2023_AEM23, Ray2026}. The SkyTEM test data are available at \url{https://github.com/GeoscienceAustralia/HiQGA.jl/}. The coordinate descent routine used is from the Lasso.jl package available at \url{https://github.com/JuliaStats/Lasso.jl}.

\bibliographystyle{rasti}
\bibliography{library}

\appendix
\section{Coordinate descent and soft thresholding for the lasso}\label{sec:whycd}
For the lasso problem of the form
\begin{align}
    \argmin_{\mathbf{m}} \phi({\mathbf{m}})
    &=
        \frac{1}{2}
        \left\|
            \mathbf{y} - \mathbf{A}\mathbf{m}
        \right\|_2^2
        +
        \lambda^2
        \left\|
            \bm\Lambda\mathbf{m}
        \right\|_1, \label{eq:x_lasso_problem_app}\\
  	\phi({\mathbf{m}}) &=\sum_{i=1}^{n_d}\frac{1}{2}\left(y_i - \sum_{j=1}^{p}A_{ij}m_j\right)^2 + \lambda^2\sum_{j=1}^{p}\Lambda_j|m_j|.
\end{align}
Writing the $i^{th}$ error \textit{without} contributions from the $m_k$ component as
\begin{equation}
r_{ik} = y_i - \sum_{\substack{j=1 \\ j \neq k}}^p A_{ij}m_j, \label{eqn:almost_residual}
\end{equation}
we can rewrite the lasso objective, including the dropped term, as, 
\begin{equation}
\phi({\mathbf{m}}) =\sum_{i=1}^{n_d}\frac{1}{2}\left(r_{ik} - A_{ik}m_k\right)^2 + \lambda^2\sum_{j=1}^{p}\Lambda_j|m_j|.
\end{equation}
Deriving now with respect to $m_k$: 
\begin{equation}
\pdv {\phi}{m_k} = -\sum_{i=1}^{n_d}A_{ik}\left(r_{ik} - A_{ik}m_k\right) + \lambda^2\Lambda_k\mathrm{sign}(m_k),
\end{equation}
we note that the above expression is valid everywhere \textit{except} exactly at $m_k=0$, without explicitly invoking mathematical subgradients. Expanding the first term,
\begin{equation}
\pdv {\phi}{m_k} = \sum_{i=1}^{n_d}A_{ik}^2 m_k - \sum_{i=1}^{n_d}A_{ik}r_{ik} + \lambda^2\Lambda_k\mathrm{sign}(m_k). \label{eqn:gradlasso}
\end{equation}
Unless each row of column $k$ is zero, $\sum_{i=1}^{n_d}A_{ik}^2 > 0$. For convenience, we write 
\begin{equation}
a_k = \sum_{i=1}^{n_d}A_{ik}^2, \qquad c_k =  \sum_{i=1}^{n_d}A_{ik}r_{ik}. \label{eq:lasso_parts}
\end{equation}
Note the similarity of $c_k$ to the ordinary least squares misfit gradient for model component $k$. Writing in this form allows us to write, for optimality, setting the gradient~\eqref{eqn:gradlasso} equal to zero:
\begin{align}
a_k m_k -c_k &= -\lambda^2\Lambda_k\mathrm{sign}(m_k),\\
m_k &= \frac{c_k -\lambda^2\Lambda_k\mathrm{sign}(m_k)}{a_k}.\label{eq:optimal_lasso}
\end{align}
Now we know that for $m_k<0$, since $a_k>0$, that the numerator of \eqref{eq:optimal_lasso} must be negative. Therefore, $c_k < -\lambda^2\Lambda_k$. Similarly, when $m_k>0$, we must have $c_k > \lambda^2\Lambda_k$. Consequently, we can write
\begin{equation}
m_k = 
\begin{cases}
\frac{c_k +\lambda^2\Lambda_k}{a_k}, & c_k < -\lambda^2\Lambda_k, \\[1.0ex]
\frac{c_k -\lambda^2\Lambda_k}{a_k}, & c_k > +\lambda^2\Lambda_k, \\[1.0ex]
0, & \mathrm{else},
\end{cases}
    \label{eq:softthresh_cases}
\end{equation}
where we have tacitly not defined a value for the derivative of the absolute value function in the region $-\lambda^2\Lambda_k < c_k < +\lambda^2\Lambda_k$, since we know that at this point, $m_k = 0$. For a more rigorous argument, this is in accordance with the theory of subdifferentials, since the conventional `gradient' of $|m|$ at 0 lies in $[-1,1]$ \citep[see e.g.,][]{rockafellar1997convex}. The cases in \eqref{eq:softthresh_cases} are summarised through the soft threshold function \eqref{eq:soft_threshold_scalar}:
\begin{equation}
S(m, l) = \mathrm{sign}(m)\max\{|m|-l, 0\},
\end{equation} 
or 
\begin{equation}
m_k = S\left(\frac {c_k}{a_k}, \frac{\lambda^2\Lambda_k}{a_k}\right). \label{eq:lasso_componentwise}
\end{equation}
A basic coordinate descent algorithm will individually cycle through all `coordinates' $k$ of  $\mathbf m$ and assign them values based on the soft thresholding operation in \eqref{eq:lasso_componentwise}. Values with weight $\Lambda_k=0$ are subject only to changes in concert with least squares reduction in misfit with no regularisation. For other terms, larger values of $\lambda^2$ shrink $m_k$ exactly to zero.

The logic behind the exact shrinkage to zero lies in the fact that $c_k$ in \eqref{eq:lasso_parts} is the $k^{th}$ component of the data misfit gradient for $m_k$ set to zero (since it is dropped from the residual in \eqref{eqn:almost_residual}). If this gradient component is small when compared to the threshold in \eqref{eq:softthresh_cases}, $m_k$ does not contribute much to the data misfit, hence it can be set to zero. Efficient ``pathwise'' coordinate descent algorithms \citep{friedman2010regularization} may interleave through multiple values of $\lambda^2$ without finishing a full sweep through all coordinates. One important special case to keep in mind is when  $\mathbf {A = I}$, as this leads to simple thresholding independent of other variables: 
\begin{equation}
m_k = S\left(y_k, \lambda^2\Lambda_k\right). \label{eq:lasso_componentwise_independent}
\end{equation}

%%%%%%%%%%%%%%%%%%%%%%%%%%%%%%%%%%%%%%%%%%%%%%%%%%

% Don't change these lines
\bsp	% typesetting comment
\label{lastpage}
\end{document}